\PassOptionsToPackage{table,xcdraw}{xcolor}
\documentclass[letterpaper]{article} 
\usepackage{aaai2026}  
\usepackage{times}  
\usepackage{helvet}  
\usepackage{courier}  
\usepackage[hyphens]{url}  
\usepackage{graphicx} 
\usepackage{natbib}  
\usepackage{caption} 
\usepackage{xcolor}
\usepackage{times}
\usepackage{epsfig}
\usepackage{graphicx}
\usepackage{subcaption}
\usepackage{amsmath}
\usepackage{amssymb}
\usepackage{textcomp}
\usepackage[linesnumbered,ruled]{algorithm2e}
\usepackage{multirow}
\usepackage{pifont}
\usepackage{overpic}
\usepackage{makecell}
\usepackage{arydshln}
\usepackage{newfloat}
\usepackage{listings}

\DeclareCaptionStyle{ruled}{labelfont=normalfont,labelsep=colon,strut=off} 
\title{\includegraphics[height=10pt]{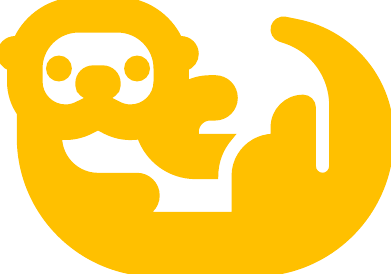} Otter: Mitigating Background Distractions of Wide-Angle Few-Shot Action Recognition with Enhanced RWKV}
\author{
Wenbo Huang\textsuperscript{\rm 1~\rm2},
Jinghui Zhang\textsuperscript{\rm 1}\thanks{Corresponding author.},
Zhenghao Chen\textsuperscript{\rm 3},
Guang Li\textsuperscript{\rm 4},
Lei Zhang\textsuperscript{\rm 5}\footnotemark[1],
Yang Cao\textsuperscript{\rm 2},
Fang Dong\textsuperscript{\rm 1},
Takahiro Ogawa\textsuperscript{\rm 4},
Miki Haseyama\textsuperscript{\rm 4}
}
\affiliations{
\textsuperscript{\rm 1}Southeast University, Nanjing 21189, Jiangsu, China; \\
\textsuperscript{\rm 2}Institute of Science Tokyo, Tokyo 152-8550, Japan; \\
\textsuperscript{\rm 3}The University of Newcastle, Callaghan, NSW 2308, Australia; \\
\textsuperscript{\rm 4}Hokkaido University, Sapporo 060-0808, Hokkaido, Japan; \\ \textsuperscript{\rm 5}Nanjing Normal University, Nanjing 210023, Jiangsu, China.

wenbohuang1002@outlook.com, \{jhzhang,~fong\}@seu.edu.cn,     zhenghao.chen@newcastle.edu.au, \{guang,~ogawa,~mhaseyama\}@lmd.ist.hokudai.ac.jp, cao@c.titech.ac.jp, leizhang@njnu.edu.cn
}

\usepackage{bibentry}

\begin{document}

\maketitle

\begin{abstract}
	Wide-angle videos in few-shot action recognition~(FSAR) effectively express actions within specific scenarios. However, without a global understanding of both subjects and background, recognizing actions in such samples remains challenging because of the background distractions. Receptance Weighted Key Value~(RWKV), which learns interaction between various dimensions, shows promise for global modeling. While directly applying RWKV to wide-angle FSAR may fail to highlight subjects due to excessive background information. Additionally, temporal relation degraded by frames with similar backgrounds is difficult to reconstruct, further impacting performance. Therefore, we design the Comp\underline{\textbf{O}}und Segmen\underline{\textbf{T}}ation and \underline{\textbf{T}}emporal R\underline{\textbf{E}}constructing \underline{\textbf{R}}WKV~(\textbf{Otter}). Specifically, the Compound Segmentation Module~(CSM) is devised to segment and emphasize key patches in each frame, effectively highlighting subjects against background information. The Temporal Reconstruction Module~(TRM) is incorporated into the temporal-enhanced prototype construction to enable bidirectional scanning, allowing better reconstruct temporal relation. Furthermore, a regular prototype is combined with the temporal-enhanced prototype to simultaneously enhance subject emphasis and temporal modeling, improving wide-angle FSAR performance. Extensive experiments on benchmarks such as SSv2, Kinetics, UCF101, and HMDB51 demonstrate that Otter achieves state-of-the-art performance. Extra evaluation on the VideoBadminton dataset further validates the superiority of Otter in wide-angle FSAR.
\end{abstract}

\begin{links}
	\link{Code}{https://github.com/wenbohuang1002/Otter}
\end{links}
\section{Introduction}
\label{sec: intro}
\indent The difficulties of video collection and labeling complicates traditional data-driven training based on fully labeled datasets. Fortunately, few-shot action recognition~(FSAR) improves learning efficiency and reduces the labeling dependency by classifying unseen actions from extremely few video samples. Therefore, FSAR has diverse real-world applications, including health monitoring and motion analysis~\cite{yan2023feature,wang2023openoccupancy}. However,  recognizing similar actions under regular viewpoint is a non-trivial problem in FSAR. For instance, distinguishing “indoor climbing” and “construction working” is challenging, as subjects exhibit similar actions against a wall. To mitigate this issue, wide-angle videos provide contextual background, such as a “climbing wall” or a “construction site”,  expressing actions within specific scenarios more accurately. According to established definitions~\cite{lai2021correcting,zhang2025madcow}, wide-angle videos with a greater field of view~(FoV) are widespread\footnote{This work adopts the widely accepted definition of wide-angle as FoV exceeding 80$^{\circ}$.}. FoV estimation~\cite{lee2021ctrl,hold2023perceptual} on popular FSAR benchmarks further reveals that approximately 35\% of samples per dataset fall into this category, yet remain unexplored. \\
\begin{figure}[t]
	\centering
	\includegraphics[width=0.46\textwidth]{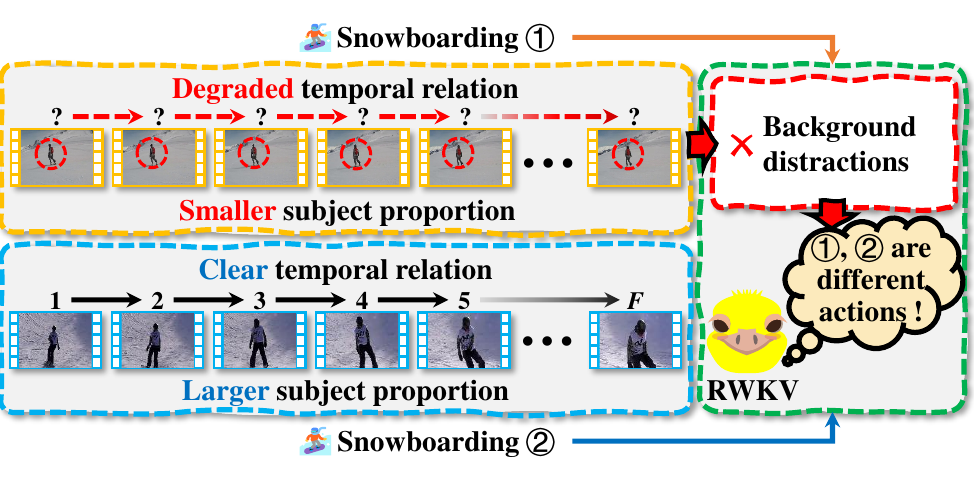} 
	\caption{Smaller subject proportion~(red circles) and degraded temporal relation~(red dotted lines) both contribute to background distractions in wide-angle FSAR. As a result, wide-angle samples are more challenging to recognize compared with regular samples.}
	\label{fig: wide}
\end{figure}
\indent On the other hand, effectively modeling wide-angle videos remains a critical issue due to the difficulty of accurately interpreting both subjects and background content. Recent success in recurrent model-based architectures has led to methods such as Receptance Weighted Key Value~(RWKV)~\cite{peng2023rwkv,peng2024eagle}, which demonstrate strong performance in global modeling across various tasks by enabling token interaction through linear interpolation, thereby expanding the receptive field and efficiently capturing subject–background dependencies.\\
\indent To seamlessly apply RWKV in wide-angle FSAR, two key challenges remain, primarily due to background distractions, as illustrated in \figurename~\ref{fig: wide}. \emph{Challenge 1: Lack of primary subject highlighting in RWKV.} As shown in the “snowboarding” examples, the primary subject occupies a smaller proportion in wide-angle frames. When RWKV is directly applied for global feature extraction, it tends to capture massive secondary background information “snow” rather than the primary subject “athlete”. Since the background serves as contextual information while the subject is crucial for determining feature representation, this reversal of primary and secondary information may lead to potential misclassification. \emph{Challenge 2: Absence of temporal relation reconstruction in RWKV.} Temporal relation plays a significant role in FSAR, primarily in perceiving action direction and aligning frames. From the “snowboarding” example, we observe that abundant background information in similar frames obscures the evolution of primary subject “athlete”, causing temporal relation degraded in wide-angle samples. However, RWKV focuses on global modeling but lacks the capability to reconstruct temporal relation, increasing the difficulty of recognizing wide-angle samples. \\
\indent Although current attempts achieve promising results~\cite{fu2020depth,wang2023molo,perrett2021temporal,huang2024soap,wang2022hybrid,xing2023revisiting}, few works address the two aforementioned challenges simultaneously. Therefore, we propose the Comp\underline{\textbf{O}}und Segmen\underline{\textbf{T}}ation and \underline{\textbf{T}}emporal R\underline{\textbf{E}}constructing \underline{\textbf{R}}WKV~(\textbf{Otter}), which highlights subjects and restores temporal relations in wide-angle FSAR. To be specific, we devise the Compound Segmentation Module~(CSM) to adaptively segment each frame into patches and highlight the subject before feature extraction. This enables RWKV to focus on the subject rather than being overwhelmed by secondary background information. We further design the Temporal Reconstruction Module (TRM), integrated into temporal-enhanced prototype construction to perform bidirectional feature scanning across frames, enabling RWKV to reconstruct temporal relations degraded in wide-angle videos. Additionally, we combine a regular prototype with a temporal-enhanced prototype to simultaneously achieve subject highlighting and temporal relation reconstruction. This strategy significantly improves the performance of wide-angle FSAR.\\
\indent To the best of our knowledge, the proposed Otter is the first attempt of utilizing RWKV for wide-angle FSAR. The core contribution is threefold.
\begin{itemize}
	\item The CSM is introduced to highlight the primary subject in RWKV. It segments each frame into multiple patches, learns adaptive weights from each patch to highlight the subject, and then reassembles the patches in their original positions. This process enables more effective detection of inconspicuous subjects in wide-angle FSAR.
	\item The TRM is designed to reconstruct temporal relations in RWKV. It performs bidirectional scanning of frame features and reconstructs the temporal relation via a weighted average of the scanning results for the temporal-enhanced prototype. This module mitigates temporal relation degradation in wide-angle FSAR.
	\item The state-of-the-art (SOTA) performance achieved by Otter is validated through extensive experiments on prominent FSAR benchmarks, including SSv2, Kinetics, UCF101, and HMDB51. Additional analyses on wide-angle VideoBadminton dataset emphasize superiority of Otter, particularly in wide-angle FSAR.
\end{itemize}
\section{Related works}
\label{sec: related}
\subsection{Few-Shot Learning}
\label{few-shot}
\indent Few-shot learning, which aims to classify unseen classes using extremely limited samples, is a crucial area in the deep learning community~\cite{fei2006one}. It encompasses three main paradigms: augmentation, optimization, and metric-based. Augmentation-based methods~\cite{hariharan2017low,wang2018low,zhang2018metagan,chen2019image,li2020adversarial} address data scarcity by generating synthetic samples to augment the training set. In contrast, optimization-based methods~\cite{finn2017model,ravi2017optimization,rusu2018meta,jamal2019task,rajeswaran2019meta} focus on modifying the optimization process to enable efficient fine-tuning with few samples. Among these approaches, the metric-based paradigm~\cite{snell2017prototypical,oreshkin2018tadam,sung2018learning,hao2019collect,wang2020cooperative} is the most widely adopted in practical applications due to its simplicity and effectiveness. Specifically, these methods construct class prototypes and perform classification by the similarity between query features and class prototypes using learnable metrics.
\begin{figure*}[t]
	\centering
	\includegraphics[width=0.99\textwidth]{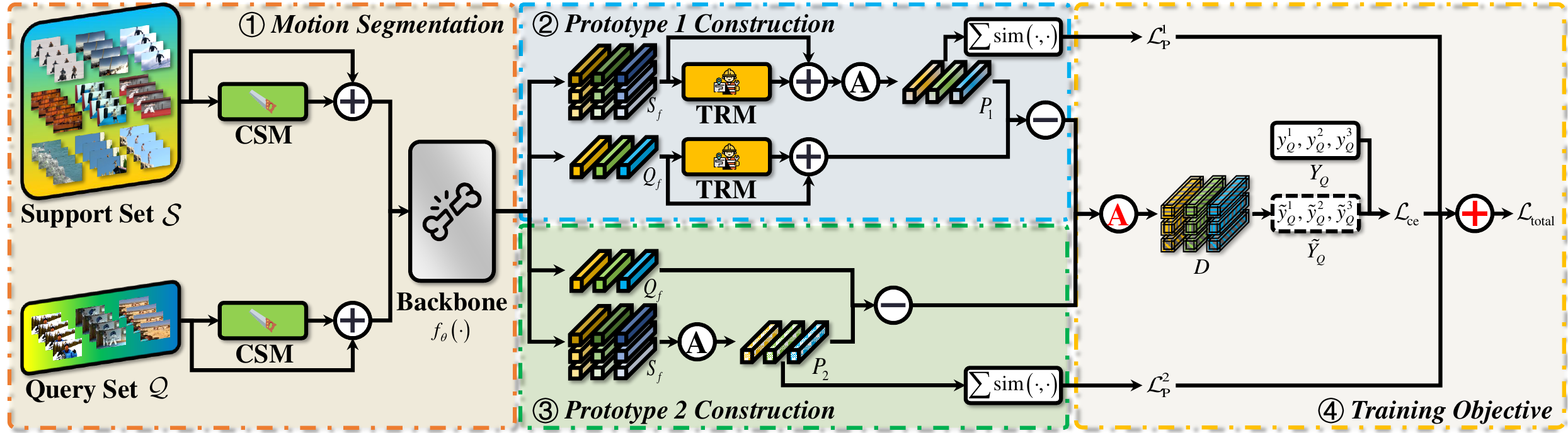} 
	\caption{The overall architecture of the Otter. Main components CSM and TRM are specified combination of core units~(\S~\ref{core units}). To be specific, \textcircled{\scriptsize{\textbf{1}}} Motion Segmentation with CSM and backbone~(\S~\ref{MS}). \textcircled{\scriptsize{\textbf{2}}} Prototype 1 Construction with TRM for reconstructing temporal relation~(\S~\ref{P1}). \textcircled{\scriptsize{\textbf{3}}} Prototype 2 Construction with regular prototype~(\S~\ref{P1}). \textcircled{\scriptsize{\textbf{4}}} Training Objective $\mathcal{L}_{\text{total}} $ is the loss combination of cross-entropy loss $\mathcal{L}_{\text{ce}} $, $\mathcal{L}^{1}_{\text{P}} $ from \textcircled{\scriptsize{\textbf{2}}}, and $\mathcal{L}^{2}_{\text{P}} $ from \textcircled{\scriptsize{\textbf{3}}}~(\S~\ref{training objective}). Notion \textcircled{\scriptsize{\textbf{A}}}/\textcircled{\scriptsize{\textcolor{red}{\textbf{A}}}}:  averaging/weighted averaging. \textcircled{\scriptsize{\textbf{+}}}/\textcircled{\scriptsize{\textcolor{red}{\textbf{+}}}} : element-wise plus/weighted element-wise plus.}
	\label{fig: overall arch}
\end{figure*}
\subsection{Few-Shot Action Recognition}
\label{fsar}
\indent Metric-based meta-learning is the mainstream paradigm in FSAR due to its simplicity and effectiveness. This approach embeds support features into class prototypes to represent various classes. Most methods rely on temporal alignment to match queries with prototypes. For example, the dynamic time warping~(DTW) algorithm is used in OTAM for similarity calculation~\cite{cao2020few}. Subsequent works, including ITANet~\cite{zhang2021learning}, T$^2$AN~\cite{li2022ta2n}, and STRM~\cite{thatipelli2022spatio}, further optimize temporal alignment. To focus more on local features, TRX~\cite{perrett2021temporal}, HyRSM~\cite{wang2022hybrid}, SloshNet~\cite{xing2023boosting}, SA-CT~\cite{zhang2023importance}, and Manta~\cite{huang2025manta} employ fine-grained or multi-scale modeling. Additionally, models are enhanced with supplementary information such as depth~\cite{fu2020depth}, optical flow~\cite{wanyan2023active}, and motion cues~\cite{wang2023molo,wu2022motion,huang2024soap}. Despite achieving satisfactory performance, They are unable to address challenges in wide-angle FSAR simultaneously.
\subsection{RWKV Model}
\label{rwkv}
\indent The RWKV model is initially proposed for natural language processing~(NLP)~\cite{peng2023rwkv,peng2024eagle}, combining the parallel processing capabilities of Transformers with the linear complexity of RNNs. This fusion enables RWKV to achieve efficient global modeling with reduced memory usage and accelerated inference speed following data-driven training. Building on this foundation, the vision-RWKV~(VRWKV) model is developed for computer vision tasks and has demonstrated notable success~\cite{duan2024vision}. Additionally, numerous studies have explored integrating RWKV with Diffusion or CLIP, achieving remarkable results in various domains~\cite{fei2024diffusion,gu2024rwkv,he2024pointrwkv,yuan2024mamba}. However, the potential of RWKV in wide-angle FSAR remains unexplored.
\section{Methodology}
\label{sec: method}
\subsection{Problem Definition} 
\label{definition}
Following settings in previous literature~\cite{cao2020few,perrett2021temporal}, three parts including training set $\mathcal{D}_\text{train}$, validation set $\mathcal{D}_\text{val}$, and testing set $\mathcal{D}_\text{test}$ without overlap~($\mathcal{D}_\text{train}\cap \mathcal{D}_\text{val}\cap \mathcal{D}_\text{test}=\varnothing $) are divided from datasets. Each part is further split into two non-overlapping sets including support~$\mathcal{S}$ with at least one labeled sample of each class and query~$\mathcal{Q}$ with all unlabeled samples~($\mathcal{S}\cap \mathcal{Q}=\varnothing $).  The aim of FSAR is to classify samples from $\mathcal{Q}$ into one class of $\mathcal{S}$. A large number of few-shot tasks are randomly selected and combined from $\mathcal{D}_\text{train}$. We define few-shot setting as $N$-way $K$-shot from  $\mathcal{S}$ with $N$ classes, $K$ samples in each class.\\
\indent Successive $F$ frames are uniformly extracted from a video each time. The $k^\text{th}$ ($k = 1, \cdots, K$) sample of the $n^\text{th}$ ($n = 1, \cdots, N$) class of $\mathcal{S}$ is defined as $S^{n,k}$ and randomly selected sample from $\mathcal{Q}$ is denoted as $Q^{r}$ ($r \in \mathbb{Z}^+$).
\begin{equation}
	\begin{aligned}
		S^{n,k}&=\left[ s_{1}^{n,k},\dots ,s_{F}^{n,k} \right] \in \mathbb{R} ^{F\times C\times H\times W}, \\
		Q^{\gamma}&=\left[ q_{1}^{\gamma},\dots ,q_{F}^{\gamma} \right] \in \mathbb{R} ^{F\times C\times H\times W},
	\end{aligned}
	\label{eq1}
\end{equation} 
in which $F$, $C$, $H$, and $W$ represent frames, channels, height, and width, respectively.
\subsection{Overall Architecture}
\label{overall}
\indent We demonstrate the overall architecture of Otter via a simple 3-way 3-shot example in \figurename~\ref{fig: overall arch}. The following two main components of Otter are built from specific combinations of core units~(\S~\ref{core units}). At the first stage of motion segmentation, CSM works for highlighting subjects before feature extracting via backbone~(\S~\ref{MS}). TRM is introduced in the second stage of prototype 1~(temporal-enhanced) construction, reconstructing the temporal relation~(\S~\ref{P1}). Prototype 2~(regular) construction is the third stage, retaining subject emphasis~(\S~\ref{P1}). Finally, distances calculated from weighted average of two prototypes are employed in cross-entropy loss~$\mathcal{L}_\text{ce}$. In order to further distinguish class prototypes, the prototype similarities serve as $\mathcal{L}_\text{p}^1$ and $\mathcal{L}_\text{p}^2$. The weighted combination of three loss including $\mathcal{L}_\text{ce}$,  $\mathcal{L}_\text{p}^1$ and $\mathcal{L}_\text{p}^2$ is the training objective $\mathcal{L}_\text{total}$~(\S~\ref{training objective}).
\subsection{Core Units}
\label{core units}
\begin{figure}[t]
	\centering
	\subcaptionbox{Spatial Mixing\label{fig: spatial_mix}}{\includegraphics[width=0.145\textwidth]{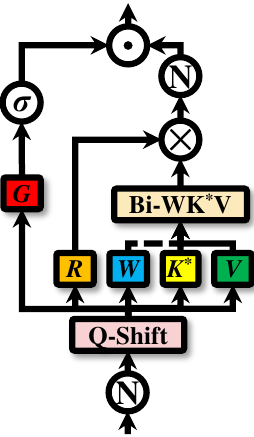}}
	\subcaptionbox{Time Mixing\label{fig: time_mix}}{\includegraphics[width=0.145\textwidth]{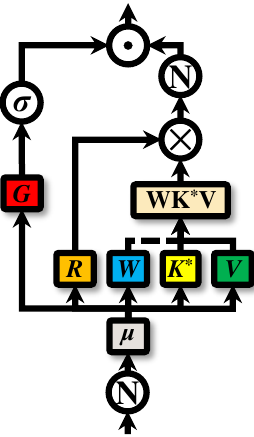}}
	\subcaptionbox{Channel Mixing\label{fig: channel_mix}}{\includegraphics[width=0.145\textwidth]{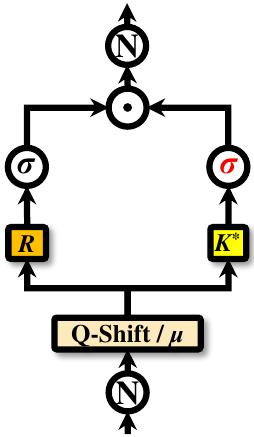}}
	\caption{Core units of RWKV. \textcircled{\scriptsize{\textbf{N}}}: normalization, \textcircled{\scriptsize{\textbf{$\times$}}}/\textcircled{\scriptsize{\textbf{$\cdot$}}}: matrix/element-wise multiplication, \textcircled{\scriptsize{\textbf{$\sigma$}}}: activation function.}
	\label{fig: rwkv}
\end{figure}
\indent In order to simplify equation writing, we use wildcard symbol $\vartriangle$. Self-attention can be simulated through five tensors: receptance $R$, weight $W$, key $K^{*}$, value $V$, and gate $G$. To handle spatial, temporal, and channel-wise features, we design three core units: Spatial Mixing, Temporal Mixing, and Channel Mixing, inspired by the architecture of RWKV-5/6. The main components, CSM and TRM, are specific combinations of these core units,  for subject highlighting and temporal relation reconstruction in wide-angle FSAR.\\
\indent To be specific, Spatial Mixing (\figurename~\ref{fig: spatial_mix}) is designed to aggregate features from different spatial locations. Let $r_t$, $k^{}_t$, $v_t$, and $g_t$ denote the $t^\text{th}$ features of $R$, $K^{*}$, $V$, and $G$, respectively. This design allows the model to capture dependencies across different regions of the image, thereby enhancing its ability to model global spatial features.
\begin{equation}
	\begin{aligned}
		\vartriangle _t &= W_{\vartriangle} \cdot \mathrm{Q}\text{-}\mathrm{Shift}_{\vartriangle}\left( x \right) 
		\\
		&= W_{\vartriangle} \cdot \left[ x + \left( 1 - \mu _{\vartriangle} \right) \odot x^{\prime} \right], \forall{\vartriangle} \in \left\{ r, k^{*}, v, g \right\}, 
		\\
		x^{\prime}_{\left[ h^{\prime}, w^{\prime} \right]} &= \mathrm{Concat}\left( x_{\left[ h^{\prime}-1, w^{\prime}, 0:C/4 \right]}, x_{\left[ h^{\prime}+1, w^{\prime}, C/4:C/2 \right]}, \right. \\
		&\quad \left. x_{\left[ h^{\prime}, w^{\prime}-1, C/2:3C/4 \right]}, x_{\left[ h^{\prime}, w^{\prime}+1, 3C/4:C \right]} \right),
	\end{aligned}
	\label{eq2}
\end{equation}
where $\mu$ is a learnable vector for the calculation of $R$, $K^{*}$, and $V$ while $\mathrm{Concat}\left( \cdot \right) $ means concatenate operation. “:” separates the start and end index. Row and column index of $x$ are denoted by $h^{\prime}$ and $w^{\prime}$. Then attention result $\left(wk^{*}v\right)_t$ is calculated according to the following definition.
\begin{equation}
	\begin{aligned}
		\left(wk^{*}v\right)_t&=\mathrm{Bi}\text{-}\mathrm{WK^{*}V}\left( K^{*},V \right)_t 
		\\
		&=\frac{\sum\nolimits_{i=0,i\ne t}^{t-1}{e^{-\left( \left| t-i \right|-1 \right) \cdot w+k^{*}_i}v_i+e^{u+k^{*}_t}v_t}}{\sum\nolimits_{i=0,i\ne t}^{t-1}{e^{-\left( \left| t-i \right|-1 \right)  \cdot w+k^{*}_i}+e^{u+k^{*}_t}}},
	\end{aligned}
	\label{eq3}
\end{equation}
$W$ is determined by vector $w$. After combining with $r_t$ and $g_t$, the $o^\text{th}$ feature of output $O$ can be calculated as
\begin{equation}
	\begin{aligned}
		o_t=\sigma \left( g_t \right) \odot \mathrm{Norm}\left( r_t\otimes \left(wk^{*}v\right)_t \right), 
	\end{aligned}
	\label{eq4}
\end{equation}
in which $\sigma \left( \cdot \right)$ denotes activation function while $\mathrm{Norm}\left( \cdot \right)$ represents normalization. \\
\indent As illustrated \figurename~\ref{fig: time_mix}, we observe that the main discrepancies between Time Mixing and Spatial Mixing are $\vartriangle _t$ and $\mathrm{WK^{*}V}\left( \cdot \right)$. The former one can be defined as
\begin{equation}
	\begin{aligned}
		\vartriangle _t = W_{\vartriangle} \cdot \left[ x_t + \left( 1 - \mu _{\vartriangle} \right) \odot x_t-1 \right], \forall{\vartriangle} \in \left\{ r, k^{*}, v, g \right\}, 
	\end{aligned}
	\label{eq5}
\end{equation}
while the latter can be written as
\begin{equation}
	\begin{aligned}
		\left(wk^{*}v\right)_t&=\mathrm{WK^{*}V}\left( K^{*},V \right)_t 
		\\
		&=\frac{\sum\nolimits_{i=0,i\ne t}^{t-1}{e^{-\left( t-i-1 \right) \cdot w+k^{*}_i}v_i+e^{u+k^{*}_t}v_t}}{\sum\nolimits_{i=0,i\ne t}^{t-1}{e^{-\left(t-i-1 \right)  \cdot w+k^{*}_i}+e^{u+k^{*}_t}}}, 
	\end{aligned}
	\label{eq6}
\end{equation}
After achieving $O$ with the same way, the combination of current and past states enable long-term modeling. \\
\indent In order to capture dependencies between multiple dimensions of input, Channel Mixing~(\figurename~\ref{fig: channel_mix}) mixes information from various channels by $R$ and $V$, as
\begin{equation}
	\begin{aligned}
		O=\sigma _r\left( R \right) \odot \sigma _v\left( V \right) .
	\end{aligned}
	\label{eq7}
\end{equation}
$\sigma _r\left( \cdot \right)$ and $\sigma _v\left( \cdot \right)$ means two difference kinds of activation function applied for $R$ and $V$.
\subsection{Motion Segmentation}
\label{MS}
\subsubsection{Compound Segmentation Module~(CSM)}
\begin{figure}[t]
	\centering
	\includegraphics[width=0.35\textwidth]{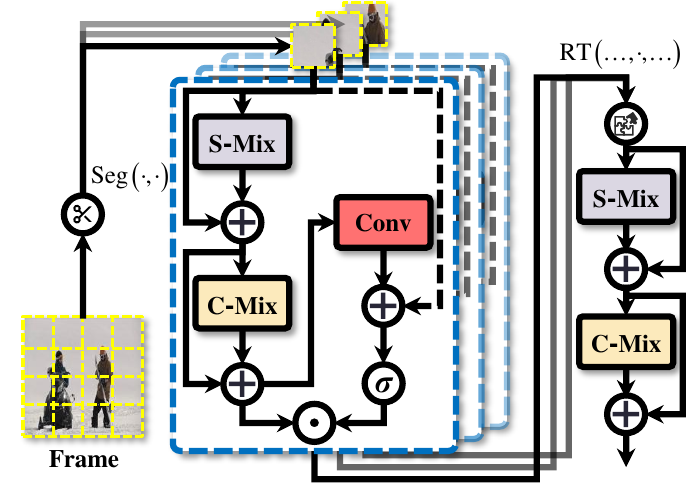} 
	\caption{The structure of Compound Segmentation Module~(CSM).}
	\label{fig: csm}
\end{figure}
\indent As demonstrated in \figurename~\ref{fig: csm}, each frame is segmented into $HW/p^2$ patches with $\mathrm{Seg}\left( \cdot, \cdot \right)$. Using random frames $s, q \in \mathbb{R} ^{C\times H\times W}$ from $S^{n,k}, Q^{r}$ as simple examples.
\begin{equation}
	\begin{aligned}
		\vartriangle ^p=\mathrm{Seg}\left( \vartriangle ,p \right) \in \mathbb{R} ^{C\times p\times p},\forall{\vartriangle} \in \left\{ s,q \right\}.
	\end{aligned}
	\label{eq8}
\end{equation} 
$H$ and $W$ must be divisible by $p$. The operations of Spatial Mixing, Time Mixing, and Channel Mixing can be written as $\mathrm{S}\text{-}\mathrm{Mix}\left( \cdot \right)$, $\mathrm{T}\text{-}\mathrm{Mix}\left( \cdot \right)$, and $\mathrm{C}\text{-}\mathrm{Mix}\left( \cdot \right)$, respectively. The output $\vartriangle ^{\alpha}$ of $\mathrm{S}\text{-}\mathrm{Mix}\left( \cdot \right)$ is connected with the input $\vartriangle ^p$ for capturing region associations of patches, as
\begin{equation}
	\begin{aligned}
		\vartriangle ^{\alpha}=\left[\mathrm{S}\text{-}\mathrm{Mix}\left( \vartriangle ^p \right) \oplus \vartriangle ^p \right]\in \mathbb{R} ^{C\times p\times p},\forall{\vartriangle} \in \left\{ s,q \right\}. 
	\end{aligned}
	\label{eq9}
\end{equation}
The activation function $\sigma \left( \cdot \right)$ in $\mathrm{S}\text{-}\mathrm{Mix}\left( \cdot \right)$ is $\mathrm{Sigmoid}\left( \cdot \right)$. Through the same method of connection with $\vartriangle ^{\alpha}$, the output $\vartriangle ^{\beta}$ of $\mathrm{T}\text{-}\mathrm{Mix}\left( \cdot \right)$ can be achieved.
\begin{equation}
	\begin{aligned}
		\vartriangle ^{\beta}=\left[ \mathrm{C}\text{-}\mathrm{Mix}\left( \vartriangle ^{\alpha} \right) \oplus \vartriangle ^{\alpha} \right] \in \mathbb{R} ^{C\times p\times p},\forall{\vartriangle} \in \left\{ s,q \right\}, 
	\end{aligned}
	\label{eq10}
\end{equation}
where the $\sigma _r\left( \cdot \right)$ and $\sigma _v\left( \cdot \right)$ of $\mathrm{C}\text{-}\mathrm{Mix}\left( \cdot \right)$ are $\mathrm{Sigmoid}\left( \cdot \right)$ and $\mathrm{Relu}\left( \cdot \right)$. Following C3-STISR~\cite{zhao2022c3}, learnable weights $lw^{\vartriangle}\in \mathbb{R} ^{C\times p\times p}$ can be achieved from $\vartriangle ^p$ and $\vartriangle ^{\beta}$ via convolution $\mathrm{Conv}\left( \cdot \right)$ and residual connection.
\begin{equation}
	\begin{aligned}
		lw^{\vartriangle}=\mathrm{Sigmoid}\left[ \mathrm{Conv}\left( \vartriangle ^{\beta} \right) \oplus \vartriangle ^p \right] , \forall{\vartriangle} \in \left\{ s,q \right\} . 
	\end{aligned}
	\label{eq11}
\end{equation}
Restoring all element-wise multiplication of $lw^{\vartriangle}$ and $\vartriangle ^{\beta}$ can highlight subject in frames. We write the corresponding operation in $\mathrm{RT}\left( \dots ,\cdot ,\dots \right) $ with the output $\dot{\vartriangle }$.
\begin{equation}
	\begin{aligned}
		\dot{\vartriangle} =\mathrm{RT}\left( \dots ,lw^{\vartriangle}\odot \vartriangle ^{\beta},\dots \right) \in \mathbb{R} ^{C\times H\times W},\forall{\vartriangle} \in \left\{ s,q \right\}. 
	\end{aligned}
	\label{eq12}
\end{equation}
According to \eqref{eq9} and \eqref{eq10}, the final outputs $\hat{\vartriangle}$~($\forall{\vartriangle} \in \left\{ s,q \right\}$) of CSM are calculated via $\mathrm{S}\text{-}\mathrm{Mix}\left( \cdot \right)$ and $\mathrm{C}\text{-}\mathrm{Mix}\left( \cdot \right)$. We place each $\hat{\vartriangle}$ in its raw position for residual connection with inputs $\hat{S}^{n,k}, \hat{Q}^{\gamma }$, thereby achieving subject highlighting.
\subsubsection{Feature Extraction}
\indent $D$-dimensional features $S^{n,k}_{f}, Q^{\gamma}_{f} \in \mathbb{R} ^{F\times D}$ are extracted by sending $\hat{S}^{n,k}, \hat{Q}^{\gamma}$ into backbone $f_{\theta}\left( \cdot \right) : \mathbb{R} ^{C\times H\times W}\mapsto \mathbb{R} ^D $.
\subsection{Prototype Construction}
\label{P1}
\subsubsection{Temporal Reconstruction Module~(TRM)}
\begin{figure}[t]
	\centering
	\includegraphics[width=0.29\textwidth]{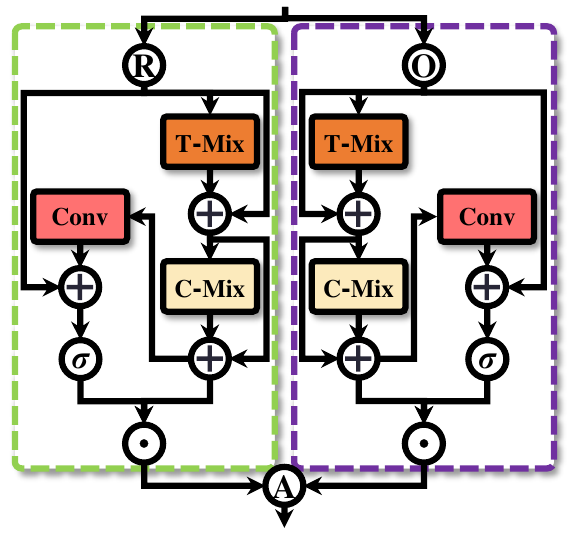} 
	\caption{The structure of Temporal Reconstruction Module~(TRM). $\normalsize{\textcircled{\scriptsize{\textbf{O}}}}$: ordered scanning. $\normalsize{\textcircled{\scriptsize{\textbf{R}}}}$: reserved scanning.}
	\label{fig: trm}
\end{figure}
\indent In order to reconstruct temporal relation, TRM illustrated in \figurename~\ref{fig: trm} has two branches for bidirectional scanning of $S^{n,k}_{f}$ and $Q^{\gamma}_{f}$. Using ordered $\mathring{\vartriangle}$ as an example, $\mathrm{T}\text{-}\mathrm{Mix}\left( \cdot \right)$ with $\mathrm{SiLU}\left( \cdot \right)$ and $\mathrm{C}\text{-}\mathrm{Mix}\left( \cdot \right)$ are applied based on \eqref{eq9} and \eqref{eq10} for long-term modeling. Learned weight $\mathring{lw}^{\vartriangle}$ can also be achieved according to \eqref{eq11}. The ordered output $\grave{\vartriangle}$ is the element-wise multiplication of $\mathring{lw}^{\vartriangle}$ and $\mathring{\vartriangle}$:
\begin{equation}
	\begin{aligned}
		\grave{\vartriangle} =\left[\mathring{lw}^{\vartriangle}\odot \mathring{\vartriangle}\right] \in \mathbb{R} ^{F\times D},\forall{\vartriangle} \in \left\{ S^{n,k}_{f},Q^{\gamma}_{f} \right\}.  
	\end{aligned}
	\label{eq13}
\end{equation}
In the same way, reversed output $\acute{\vartriangle}$ can also be achieved. The final result $\tilde{\vartriangle}$ is the average $\mathrm{Avg}\left(\cdot , \cdot \right)$ of $\grave{\vartriangle}$ and $\acute{\vartriangle}$ connected with the original input, as:
\begin{equation}
	\begin{aligned}
		\tilde{\vartriangle} =\left[ \vartriangle +\mathrm{Avg}\left( \grave{\vartriangle} ,\acute{\vartriangle} \right) \right] \in \mathbb{R} ^{F\times D},\forall{\vartriangle} \in \left\{ S^{n,k}_{f},Q^{\gamma}_{f} \right\}.  
	\end{aligned}
	\label{eq14}
\end{equation}
After the TRM, temporal relation is recovered.
\subsubsection{Prototype and Distance}
\indent $P^{n}_{1}$ is prototype of the $n^\text{th}$ support class, being achieved via average calculation of $\tilde{S}_{f}^{n,k}$:
\begin{equation}
	\begin{aligned}
		P_{1}^{n}=\small{\frac{1}{K}}\sum_{k=1}^K{\tilde{S}_{f}^{n,k}}\in \mathbb{R} ^{F\times D}.  
	\end{aligned}
	\label{eq15}
\end{equation}
The distance between $\tilde{Q}^{\gamma}_{f}$ and $P_{1}^{n}$ is $D_1$.
\begin{equation}
	\begin{aligned}
		D_1=\left\| P_{1}^{n}-\tilde{Q}_{f}^{\gamma} \right\| .  
	\end{aligned}
	\label{eq16}
\end{equation}
For further distinguishing classes of the prototype $P_1$, we apply the sum of cosine similarity function $\mathrm{Sim}\left(\cdot, \cdot \right)$ for $\mathcal{L}^{1}_{\text{P}}$:
\begin{equation}
	\begin{aligned}
		\mathcal{L} _{\mathrm{P}}^{1}=\sum_{n\ne n^{\prime}}{\mathrm{Sim}\left( P_{1}^{n},P_{1}^{n^{\prime}} \right)},\left( P_{1}^{n},P_{1}^{n^{\prime}} \right) \in P_1.  
	\end{aligned}
	\label{eq17}
\end{equation}
\indent The prototype 2 is constructed without TRM. Therefore, the $n^\text{th}$ support prototype $P^{n}_{2}$ can be computed from $S^{n,k}_{f}$. Then the corresponding distance $D_2$ between $Q^{\gamma}_{f}$ and $P_{2}^{n}$ can also be achieved. After the same cosine similarity calculation, $\mathcal{L}^{2}_{\text{P}}$ is applied for differentiating classes of $P_2$. 
\subsection{Training Objective}
\label{training objective}
\indent The distance $D$ between $n^\text{th}$ class and $Q^{\gamma}_{f}$ is the weighted mean value of $D_1$ and $D_2$ with weight $\omega$. Therefore, the predicted label $\tilde{y}_Q^j\in \tilde{Y}_Q$ of query is
\begin{equation}
	\begin{aligned}
		\tilde{y}_Q^j=\underset{n}{\mathrm{argmin}}\left( D \right), D= \sum_{i=1}^2{\omega_iD_i} .
	\end{aligned}
	\label{eq18}
\end{equation}
$\tilde{y}_Q^j$ and the ground truth $y_Q^j\in Y_Q$ are applied in cross-entropy loss $\mathcal{L}_\text{ce}$ calculation.
\begin{equation}
	\begin{aligned}
		\mathcal{L} _{\text{ce}}=-\frac{1}{N}\sum_{j=1}^N{y_Q^j\log \left( \tilde{y}_Q^j \right)}.
	\end{aligned}
	\label{eq19}
\end{equation}
The training objective $\mathcal{L}_\text{total}$ is the combination of $\mathcal{L}_\text{ce}$, $\mathcal{L}^{1}_{\text{P}}$, and $\mathcal{L}^{2}_{\text{P}}$ under weight factor $\lambda$ as:
\begin{equation}
	\begin{aligned}
		\mathcal{L} _{\text{total}}=\lambda_0 \mathcal{L} _{\text{ce}}+ \lambda_1 \mathcal{L}^{1}_{\text{P}}+ \lambda_2 \mathcal{L}^{2}_{\text{P}} ,
	\end{aligned}
	\label{eq20}
\end{equation}
\section{Experiments}
\label{sec: experiment}
\subsection{Experimental Configuration}
\begin{table*}[ht]
	\centering
	\small
	\setlength{\tabcolsep}{4.4pt}
	\begin{tabular}{l|r|r|cc|cc|cc|cc}
		\Xhline{1pt}
		\multirow{2}{*}{Methods} & \multirow{2}{*}{Reference} & \multicolumn{1}{c|}{\multirow{2}{*}{Pre-Backbone}} & \multicolumn{2}{c|}{SSv2} & \multicolumn{2}{c|}{Kinetics} & \multicolumn{2}{c|}{UCF101} & \multicolumn{2}{c}{HMDB51} \\ \cline{4-11} 
		&  & \multicolumn{1}{c|}{} & \multicolumn{1}{c|}{1-shot} & 5-shot & \multicolumn{1}{c|}{1-shot} & 5-shot & \multicolumn{1}{c|}{1-shot} & 5-shot & \multicolumn{1}{c|}{1-shot} & 5-shot \\ \hline
		STRM~\cite{thatipelli2022spatio} & CVPR'22 & ImageNet-RN50 & \multicolumn{1}{c|}{N/A} & 68.1 & \multicolumn{1}{c|}{N/A} & 86.7 & \multicolumn{1}{c|}{N/A} & 96.9 & \multicolumn{1}{c|}{N/A} & 76.3 \\
		SloshNet~\cite{xing2023revisiting} & AAAI'23 & ImageNet-RN50 & \multicolumn{1}{c|}{46.5} & 68.3 & \multicolumn{1}{c|}{N/A} & 87.0 & \multicolumn{1}{c|}{N/A} & 97.1 & \multicolumn{1}{c|}{N/A} & 77.5 \\
		SA-CT~\cite{zhang2023importance} & MM'23 & ImageNet-RN50 & \multicolumn{1}{c|}{48.9} & 69.1 & \multicolumn{1}{c|}{71.9} & 87.1 & \multicolumn{1}{c|}{85.4} & 96.3 & \multicolumn{1}{c|}{61.2} & 76.9 \\
		GCSM~\cite{yu2023multi} & MM'23 & ImageNet-RN50 & \multicolumn{1}{c|}{N/A} & N/A & \multicolumn{1}{c|}{74.2} & 88.2 & \multicolumn{1}{c|}{86.5} & 97.1 & \multicolumn{1}{c|}{61.3} & 79.3 \\
		GgHM~\cite{xing2023boosting} & ICCV'23 & ImageNet-RN50 & \multicolumn{1}{c|}{54.5} & 69.2 & \multicolumn{1}{c|}{74.9} & 87.4 & \multicolumn{1}{c|}{85.2} & 96.3 & \multicolumn{1}{c|}{61.2} & 76.9 \\ \cdashline{3-11}[1pt/1pt] 
		STRM~\cite{thatipelli2022spatio} & CVPR'22 & ImageNet-ViT & \multicolumn{1}{c|}{N/A} & 70.2 & \multicolumn{1}{c|}{N/A} & 91.2 & \multicolumn{1}{c|}{N/A} & \underline98.1 & \multicolumn{1}{c|}{N/A} & 81.3 \\
		SA-CT~\cite{zhang2023importance} & MM'23 & ImageNet-ViT & \multicolumn{1}{c|}{N/A} & 66.3 & \multicolumn{1}{c|}{N/A} & 91.2 & \multicolumn{1}{c|}{N/A} & 98.0 & \multicolumn{1}{c|}{N/A} & 81.6 \\ \cline{2-11} 
		$^{\star}$TRX~\cite{perrett2021temporal} & CVPR'21 & ImageNet-RN50 & \multicolumn{1}{c|}{53.8} & 68.8 & \multicolumn{1}{c|}{74.9} & 85.9 & \multicolumn{1}{c|}{85.7} & 96.3 & \multicolumn{1}{c|}{83.5} & 85.5 \\
		$^{\star}$HyRSM~\cite{wang2022hybrid} & CVPR'22 & ImageNet-RN50 & \multicolumn{1}{c|}{54.1} & 68.7 & \multicolumn{1}{c|}{73.5} & 86.2 & \multicolumn{1}{c|}{83.6} & 94.6 & \multicolumn{1}{c|}{80.2} & 86.1 \\
		$^{\star}$MoLo~\cite{wang2023molo} & CVPR'23 & ImageNet-RN50 & \multicolumn{1}{c|}{56.6} & 70.7 & \multicolumn{1}{c|}{74.2} & 85.7 & \multicolumn{1}{c|}{86.2} & 95.4 & \multicolumn{1}{c|}{87.3} & 86.3 \\
		$^{\star}$SOAP~\cite{huang2024soap} & MM'24 & ImageNet-RN50 & \multicolumn{1}{c|}{61.9} & 85.8 & \multicolumn{1}{c|}{86.1} & 93.8 & \multicolumn{1}{c|}{94.1} & \cellcolor[HTML]{EFEFEF}\underline{99.3} & \multicolumn{1}{c|}{86.4} & 88.4 \\
		$^{\star}$Manta~\cite{huang2025manta} & AAAI'25 & ImageNet-RN50 & \multicolumn{1}{c|}{\cellcolor[HTML]{EFEFEF}\underline{63.4}} & \cellcolor[HTML]{EFEFEF}\underline{87.4} & \multicolumn{1}{c|}{\cellcolor[HTML]{EFEFEF}\underline{87.4}} & \cellcolor[HTML]{EFEFEF}\underline{94.2} & \multicolumn{1}{c|}{\cellcolor[HTML]{EFEFEF}\underline{95.9}} & 99.2 & \multicolumn{1}{c|}{\cellcolor[HTML]{EFEFEF}\underline{86.8}} & \cellcolor[HTML]{EFEFEF}\underline{88.6} \\ \cdashline{3-11}[1pt/1pt] 
		$^{\star}$MoLo~\cite{wang2023molo} & CVPR'23 & ImageNet-ViT & \multicolumn{1}{c|}{61.1} & 71.7 & \multicolumn{1}{c|}{78.9} & 95.8 & \multicolumn{1}{c|}{88.4} & 97.6 & \multicolumn{1}{c|}{81.3} & 84.4 \\
		$^{\star}$SOAP~\cite{huang2024soap} & MM'24 & ImageNet-ViT & \multicolumn{1}{c|}{\cellcolor[HTML]{EFEFEF}\underline{66.7}} & 87.2 & \multicolumn{1}{c|}{\cellcolor[HTML]{EFEFEF}\underline{89.9}} & 95.5 & \multicolumn{1}{c|}{96.8} & \cellcolor[HTML]{EFEFEF}\underline{99.5} & \multicolumn{1}{c|}{\cellcolor[HTML]{EFEFEF}\underline{89.3}} & \cellcolor[HTML]{EFEFEF}\underline{89.8} \\
		$^{\star}$Manta~\cite{huang2025manta} & AAAI'25 & ImageNet-ViT & \multicolumn{1}{c|}{66.2} & \cellcolor[HTML]{EFEFEF}\underline{89.3} & \multicolumn{1}{c|}{88.2} & \cellcolor[HTML]{EFEFEF}\underline{96.3} & \multicolumn{1}{c|}{\cellcolor[HTML]{EFEFEF}\underline{97.2}} & \cellcolor[HTML]{EFEFEF}\underline{99.5} & \multicolumn{1}{c|}{88.9} & 88.8 \\ \cdashline{3-11}[1pt/1pt]  
		$^{\star}$MoLo~\cite{wang2023molo} & CVPR'23 & ImageNet-ViR & \multicolumn{1}{c|}{60.9} & 71.8 & \multicolumn{1}{c|}{79.1} & 95.7 & \multicolumn{1}{c|}{88.2} & 97.5 & \multicolumn{1}{c|}{81.2} & 84.6 \\
		$^{\star}$SOAP~\cite{huang2024soap} & MM'24 & ImageNet-ViR & \multicolumn{1}{c|}{66.4} & 87.1 & \multicolumn{1}{c|}{\cellcolor[HTML]{EFEFEF}\underline{89.8}} & 95.8 & \multicolumn{1}{c|}{96.6} & 99.1 & \multicolumn{1}{c|}{\cellcolor[HTML]{EFEFEF}\underline{88.8}} & \cellcolor[HTML]{EFEFEF}\underline{89.7} \\
		$^{\star}$Manta~\cite{huang2025manta} & AAAI'25 & ImageNet-ViR & \multicolumn{1}{c|}{\cellcolor[HTML]{EFEFEF}\underline{66.5}} & \cellcolor[HTML]{EFEFEF}\underline{89.2} & \multicolumn{1}{c|}{88.1} & \cellcolor[HTML]{EFEFEF}\underline{96.1} & \multicolumn{1}{c|}{\cellcolor[HTML]{EFEFEF}\underline{96.7}} & \cellcolor[HTML]{EFEFEF}\underline{99.2} & \multicolumn{1}{c|}{88.7} & 89.5 \\ \hline
		AmeFu-Net~\cite{fu2020depth} & MM'20 & ImageNet-RN50 & \multicolumn{1}{c|}{N/A} & N/A & \multicolumn{1}{c|}{74.1} & 86.8 & \multicolumn{1}{c|}{85.1} & 95.5 & \multicolumn{1}{c|}{60.2} & 75.5 \\
		MTFAN~\cite{wu2022motion} & CVPR'22 & ImageNet-RN50 & \multicolumn{1}{c|}{45.7} & 60.4 & \multicolumn{1}{c|}{74.6} & 87.4 & \multicolumn{1}{c|}{84.8} & 95.1 & \multicolumn{1}{c|}{59.0} & 74.6 \\
		AMFAR~\cite{wanyan2023active} & CVPR'23 & ImageNet-RN50 & \multicolumn{1}{c|}{61.7} & 79.5 & \multicolumn{1}{c|}{80.1} & 92.6 & \multicolumn{1}{c|}{91.2} & 99.0 & \multicolumn{1}{c|}{73.9} & 87.8 \\ \cline{2-11} 
		$^{\star}$Lite-MKD~\cite{liu2023lite} & MM'23 & ImageNet-RN50 & \multicolumn{1}{c|}{55.7} & 69.9 & \multicolumn{1}{c|}{75.0} & 87.5 & \multicolumn{1}{c|}{85.3} & 96.8 & \multicolumn{1}{c|}{66.9} & 74.7 \\ \cdashline{3-11}[1pt/1pt] 
		$^{\star}$Lite-MKD~\cite{liu2023lite} & MM'23 & ImageNet-ViT & \multicolumn{1}{c|}{59.1} & 73.6 & \multicolumn{1}{c|}{78.8} & 90.6 & \multicolumn{1}{c|}{89.6} & 98.4 & \multicolumn{1}{c|}{71.1} & 77.4 \\ \cdashline{3-11}[1pt/1pt]
		$^{\star}$Lite-MKD~\cite{liu2023lite} & MM'23 & ImageNet-ViR & \multicolumn{1}{c|}{59.1} & 73.7 & \multicolumn{1}{c|}{78.5} & 90.5 & \multicolumn{1}{c|}{89.7} & 97.9 & \multicolumn{1}{c|}{71.2} & 77.5 \\ \hline
		\rowcolor[HTML]{C0C0C0}
		Otter & Ours & ImageNet-RN50 & \multicolumn{1}{c|}{\textbf{64.7}} & \textbf{88.5} & \multicolumn{1}{c|}{\textbf{90.5}} & \textbf{96.4} & \multicolumn{1}{c|}{\textbf{96.8}} & \textbf{99.2} & \multicolumn{1}{c|}{\textbf{88.1}} & \textbf{89.8} \\ \cdashline{3-11}[1pt/1pt]
		\rowcolor[HTML]{C0C0C0} 
		Otter & Ours & ImageNet-ViT & \multicolumn{1}{c|}{\textbf{67.2}} & \textbf{89.9} & \multicolumn{1}{c|}{\textbf{91.8}} & \textbf{97.3} & \multicolumn{1}{c|}{\textbf{97.7}} & \textbf{99.4} & \multicolumn{1}{c|}{\textbf{89.9}} & \textbf{90.6} \\ \cdashline{3-11}[1pt/1pt] 
		\rowcolor[HTML]{C0C0C0} 
		Otter & Ours & ImageNet-ViR & \multicolumn{1}{c|}{\textbf{67.1}} & \textbf{89.8} & \multicolumn{1}{c|}{\textbf{91.7}} & \textbf{96.8} & \multicolumn{1}{c|}{\textbf{97.5}} & \textbf{99.3} & \multicolumn{1}{c|}{\textbf{89.5}} & \textbf{90.5} \\
		\Xhline{1pt}
	\end{tabular}
	\caption{Comparison~($\uparrow$ Acc.~\%) on ResNet-50~(ImageNet-RN50), ViT-B~(ImageNet-ViT), and VRWKV-B~(ImageNet-ViR) are separated by dashed line. \textbf{Bold texts} denotes the global best results while \underline{Underline texts} serve as the local best. From top to bottom, the whole table is divided into three parts including RGB-based, multimodal, and our Otter. In the first two parts, “$\star$” represents our implementation with the same setting. “N/A” indicates not available.}
	\label{tab: acc com}
\end{table*}
\subsubsection{Data Processing}
\indent Temporal-related SSv2~\cite{goyal2017something}, spatial-related Kinetics~\cite{carreira2017quo}, UCF101~\cite{kay2017kinetics}, and HMDB51~\cite{kuehne2011hmdb} are most frequently-used benchmark datasets for FSAR. A wide-angle dataset VideoBadminton~\cite{li2024benchmarking} is employed for evaluating real-world performance. In order to prove the effectiveness of our Otter, the sampling intervals setting of decoding videos are each 1 frame. Based on widely-used data split~\cite{zhu2018compound,cao2020few,zhang2020few}, $\mathcal{D}_\text{train}$, $\mathcal{D}_\text{val}$, and $\mathcal{D}_\text{test}$~($\mathcal{D}_\text{train}\cap \mathcal{D}_\text{val}\cap \mathcal{D}_\text{test}=\varnothing $) are divided from each dataset. Then further split of support $\mathcal{S}$ and query $\mathcal{Q}$ are executed for FSAR.\\
\indent According to TSN~\cite{wang2016temporal}, each frame are sized into $3\times 256\times 256$ while $F$ of successive frames is set to 8. $3\times 224\times 224$ random crops and horizontal flipping data augmentation is applied during training while only the center crop is utilized in testing. As an exception, horizontal flipping is absent in SSv2 because of many actions with horizontal direction such as “Pulling S from left to right\footnote{“S” means “something”.}”.
\subsubsection{Implementation Details and Evaluation Metrics}
\indent Standard 5-way 1-shot and 5-shot setting are adopted for FSAR. We select ResNet-50, ViT-B, VMamba-B, and VRWKV-B with ImageNet pre-trained weights initialization as our backbone. The dimension $D$ of features is 2048. \\
\indent The larger SSv2 are trained with 75,000 tasks while other datasets only require 10,000 tasks. SGD optimization for training is applied with initial learning rate $10^{-3}$. The $\mathcal{D}_\text{val}$ determines hyper-parameters such as distance weight~($\omega_1=\omega_2=0.5$), weight factor of loss $\lambda$~($\lambda_0=0.8, \lambda_1=\lambda_2=0.1$) and patch size~($p=56$). Average accuracy of 10,000 random tasks from $\mathcal{D}_\text{test}$ is recorded during testing stage. Experiments are most conducted on a server with two 32GB NVIDIA Tesla V100 PCIe GPUs.
\subsection{Comparison with Various Methods}
\indent We implement many methods under the same setting for fair comparison with Otter. The average accuracy~($\uparrow$ higher indicates better) is illustrated in \tablename~\ref{tab: acc com}.
\subsubsection{ResNet-50 Methods}
\indent Using SSv2 under 1-shot setting as representative results, we find that Otter outperforms the current SOTA method Manta which focuses on long sub-sequences from 63.4\% to 64.7\%. A similar improvement can also be discovered in other datasets with different shots. 
\subsubsection{ViT-B Methods}
\indent The larger model capacity makes ViT-B perform better than ResNet-50. We observe that the previous SOTA performance is achieved by SOAP or Manta. Being similar with ResNet-50, Otter reveals superior performance, surpassing previous methods.  
\subsubsection{VRWKV-B Methods}
\indent As an emerging model, VRWKV-B can efficiently extract feature form promising regions association. Compared with other backbones, we observe that the overall trend in performance has no significant changes. The proposed Otter focus on improving wide-angle samples, achieving new SOTA performance.    
\subsection{Essential Components and Factors}
\label{factor}
\subsubsection{Key Components}
\indent In order to analyze the effect of key components in Otter, we conduct experiments with only CSM, TRM, and both of them. As demonstrated in \tablename~\ref{tab: key}, we observe that CSM and TRM both improve the performance. In our design, CSM highlights subject within wide-angle frames before feature extraction. Then TRM reconstructs the degraded temporal relations. Two modules operate successively and complement each other, indicating that full Otter achieves optimal performance.
\begin{table}[t]
	\centering
	\small
	\begin{tabular}{c|c|cc|cc}
		\Xhline{1pt}
		\multirow{2}{*}{CSM} & \multirow{2}{*}{TRM} & \multicolumn{2}{c|}{SSv2} & \multicolumn{2}{c}{Kinetics} \\ \cline{3-6} 
		&  & \multicolumn{1}{c|}{1-shot} & 5-shot & \multicolumn{1}{c|}{1-shot} & 5-shot \\ \hline
		\ding{55} & \ding{55} & \multicolumn{1}{c|}{54.6} & 69.2 & \multicolumn{1}{c|}{78.1} & 85.3 \\ \hline
		\ding{51} & \ding{55} & \multicolumn{1}{c|}{61.3} & 85.6 & \multicolumn{1}{c|}{89.4} & 94.8 \\
		\ding{55} & \ding{51} & \multicolumn{1}{c|}{59.5} & 83.4 & \multicolumn{1}{c|}{87.8} & 92.7 \\ \hline
		\rowcolor[HTML]{C0C0C0}\ding{51} & \ding{51} & \multicolumn{1}{c|}{\textbf{64.7}} & \textbf{88.5} & \multicolumn{1}{c|}{\textbf{90.5}} & \textbf{96.4} \\
		\Xhline{1pt}
	\end{tabular}
	\caption{Comparison~($\uparrow$ Acc.~\%) of key components. }
	\label{tab: key}
\end{table}
\subsubsection{Patch Design in CSM}
\indent A deeper research on patch design in CSM is indicated in \tablename~\ref{tab: patch}. It is obvious that the performance is increasing with more fine grained design segmentation~(less $p$). If $p$ is further reduced to 28, the performance will have a decline. We also consider multi-scale patch configurations and observe that $p{=}56$ consistently performs better. This may be attributed to the fact that multi-scale design introduces redundant features. Therefore, we adopt $p{=}56$ with $4{\times}4$ segmentation in our patch design.
\begin{table}[t]
	\centering
	\small
	\begin{tabular}{l|cc|cc}
		\Xhline{1pt}
		\multicolumn{1}{c|}{\multirow{2}{*}{$p$}} & \multicolumn{2}{c|}{SSv2} & \multicolumn{2}{c}{Kinetics} \\ \cline{2-5} 
		\multicolumn{1}{c|}{} & \multicolumn{1}{c|}{1-shot} & 5-shot & \multicolumn{1}{c|}{1-shot} & 5-shot \\ \hline
		$p=224$ & \multicolumn{1}{c|}{62.7} & 86.4 & \multicolumn{1}{c|}{87.7} & 94.6 \\
		$p=112$ & \multicolumn{1}{c|}{63.6} & 87.1 & \multicolumn{1}{c|}{89.5} & 95.2 \\
		\rowcolor[HTML]{C0C0C0}$p=56$ & \multicolumn{1}{c|}{\textbf{64.7}} & \textbf{88.5} & \multicolumn{1}{c|}{\textbf{90.5}} & \textbf{96.4} \\
		$p=28$ & \multicolumn{1}{c|}{64.1} & 87.9 & \multicolumn{1}{c|}{90.2} & 95.8 \\ \hline
		$p\in \left\{ 28,56 \right\} $ & \multicolumn{1}{c|}{64.2} & 88.1 & \multicolumn{1}{c|}{90.1} & 96.1 \\
		$p\in \left\{ 56,112 \right\} $ & \multicolumn{1}{c|}{63.7} & 87.9 & \multicolumn{1}{c|}{89.6} & 95.8 \\
		\Xhline{1pt}
	\end{tabular}
	\caption{Comparison~($\uparrow$ Acc.~\%) of patch design in CSM.}
	\label{tab: patch}
\end{table}
\subsubsection{Direction Design in TRM}
\indent As illustrated in \tablename~\ref{tab: direction}, the experiments with unidirectional and bidirectional scanning is conducted to verify the effect of direction design in TRM. Two types of unidirectional scanning are inferior to bidirectional design. The reserved scanning~($\normalsize{\textcircled{\scriptsize{\textbf{R}}}}$) even harms the performance of ordered scanning~($\normalsize{\textcircled{\scriptsize{\textbf{O}}}}$). This may be explained by the confusion of directional related actions. Therefore, the bidirectional design is indispensable in TRM.
\begin{table}[t]
	\centering
	\small
	\begin{tabular}{c|c|cc|cc}
		\Xhline{1pt}
		\multirow{2}{*}{$\normalsize{\textcircled{\scriptsize{\textbf{O}}}}$} & \multirow{2}{*}{$\normalsize{\textcircled{\scriptsize{\textbf{R}}}}$} & \multicolumn{2}{c|}{SSv2} & \multicolumn{2}{c}{Kinetics} \\ \cline{3-6} 
		&  & \multicolumn{1}{c|}{1-shot} & 5-shot & \multicolumn{1}{c|}{1-shot} & 5-shot \\ \hline
		\ding{51} & \ding{55} & \multicolumn{1}{c|}{63.2} & 87.3 & \multicolumn{1}{c|}{89.7} & 95.7 \\
		\ding{55} & \ding{51} & \multicolumn{1}{c|}{60.6} & 85.2 & \multicolumn{1}{c|}{89.1} & 94.2 \\ \hline
		\rowcolor[HTML]{C0C0C0}\ding{51} & \ding{51} & \multicolumn{1}{c|}{\textbf{64.7}} & \textbf{88.5} & \multicolumn{1}{c|}{\textbf{90.5}} & \textbf{96.4} \\
		\Xhline{1pt}
	\end{tabular}
	\caption{Comparison~($\uparrow$ Acc.~\%) of direction design in TRM.}
	\label{tab: direction}
\end{table}
\subsection{Wide-Angle Evaluation}
\label{angle}
\subsubsection{Performance on Wide-Angle Dataset}
\indent In order to evaluate Otter on wide-angle scenario, we employ VideoBadminton dataset with all wide-sample samples for testing. Form the results in \tablename~\ref{tab: vb}, it is Otter that obviously far ahead of other methods without specific design for wide-angle samples. Owing to highlighted subject and reconstructed temporal relation, Otter mitigates background distractions. Therefore, the performance on challenging wide-angle samples is significantly improved. 
\begin{table}[ht]
	\centering
	\small
	\setlength{\tabcolsep}{4.4pt}
	\begin{tabular}{l|cc|cc}
		\Xhline{1pt}
		\multirow{2}{*}{Methods} & \multicolumn{2}{c|}{VB$\rightarrow$VB} & \multicolumn{2}{c}{KI$\rightarrow$VB} \\ \cline{2-5} 
		& \multicolumn{1}{c|}{1-shot} & 5-shot & \multicolumn{1}{c|}{1-shot} & 5-shot \\ \hline
		MoLo & \multicolumn{1}{c|}{60.2} & 64.5 & \multicolumn{1}{c|}{58.9} & 61.7 \\
		SOAP & \multicolumn{1}{c|}{63.5} & 66.9 & \multicolumn{1}{c|}{60.1} & 63.1 \\
		Manta & \multicolumn{1}{c|}{64.1} & 67.1 & \multicolumn{1}{c|}{62.1} & 65.3 \\ \hline
		\rowcolor[HTML]{C0C0C0}Otter & \multicolumn{1}{c|}{\textbf{71.2}} & \textbf{75.8} & \multicolumn{1}{c|}{\textbf{69.5}} & \textbf{72.6} \\ 
		\Xhline{1pt}
	\end{tabular}
	\caption{Comparison~($\uparrow$ Acc.~\%) with wide-angle dataset. VB$\rightarrow$VB: training and testing both on VideoBadminton, KI$\rightarrow$VB: Kinetics training while VideoBadminton testing.}
	\label{tab: vb}
\end{table}
\subsubsection{CAM Visualization}
\begin{figure}[t]
	\centering
	\includegraphics[width=0.42\textwidth]{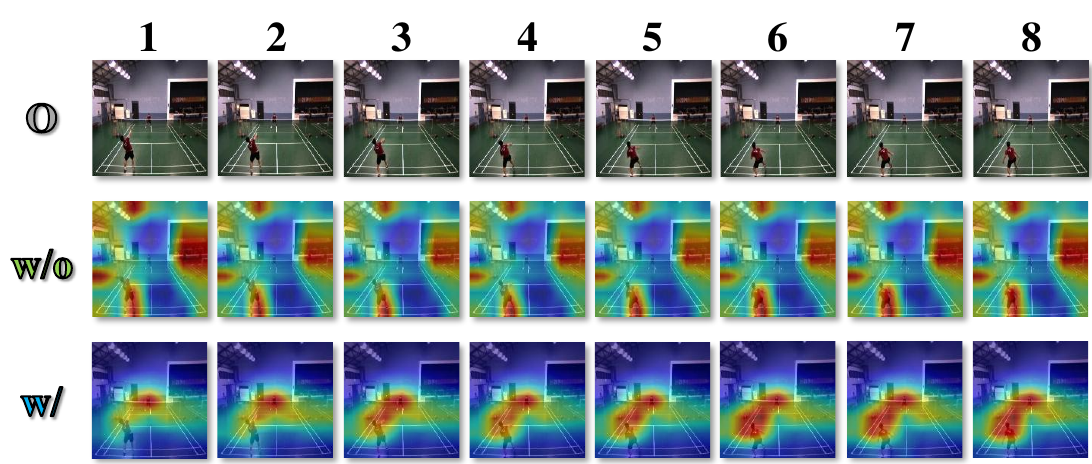} 
	\caption{CAM of “smash” from VideoBadminton. O: Original, w/: with Otter, w/o: without Otter.}
	\label{fig: background}
\end{figure}

\indent In \figurename~\ref{fig: background}, subjects are inconspicuous and similar background makes temporal relation degradation. From the CAM results without Otter, the focuses of model are mostly in the background while the subject in distance is entirely ignored. When being equipped with Otter, most of focuses are transferred to subjects and background is not completely overlooked. Compared with only focusing on the subject nearby, Otter can capture both of subjects playing badminton. These prove that Otter helps the model better understand “smash”, an action that requires interaction between two subjects, mitigating background distractions and achieving better performance in wide-angle FSAR. 
\subsubsection{Various FoV}
\begin{figure}[ht]
	\centering
	\subcaptionbox{1-shot\label{fig: wide_ssv2}}{\includegraphics[width=0.23\textwidth]{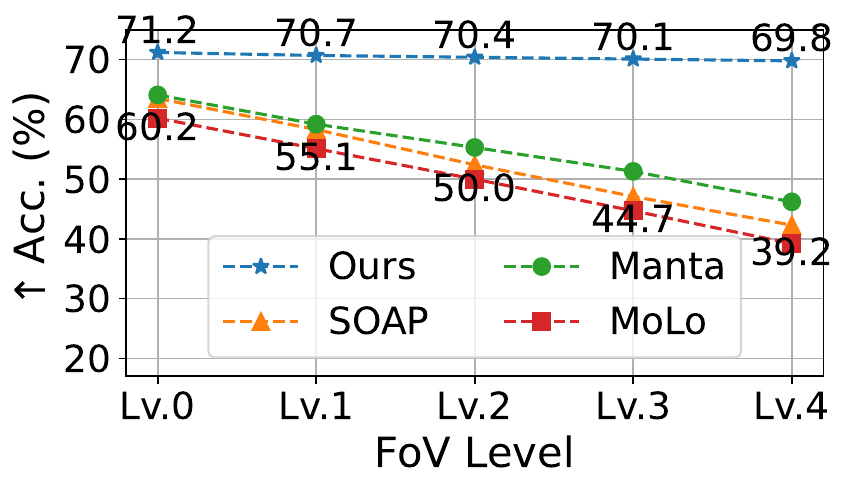}}
	\subcaptionbox{5-shot\label{fig: wide_kinetics}}{\includegraphics[width=0.23\textwidth]{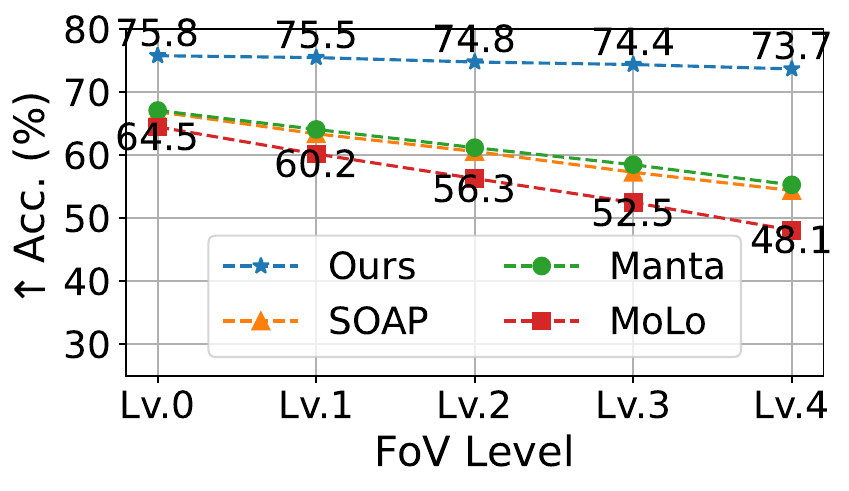}}
	\caption{Comparison~($\uparrow$~Acc.~\%) with various FoV levels.}
	\label{fig: plot_wide}
\end{figure}

\indent To rigorously evaluate Otter on wide-angle samples, frames with varying FoV are essential. Given that FoV is primarily determined by complementary metal oxide semiconductor~(CMOS) size and lens focal length~\cite{liao2023deep}, we utilized PQDiff~\cite{zhang2024continuous} for outpainting magnification~($U_\text{m}$) and introduced the distortion factor~($U_\text{d}$) in the VideoBadminton dataset to simulate diverse CMOS sizes and focal lengths. This approach results in five distinct FoV levels, with higher levels indicating a wider FoV. As indicated in \figurename~\ref{fig: plot_wide}, we observe that recent methods all have a drastic downward trend with the increasing level of FoV. Although our Otter is also negatively influenced, the downward trend is much more stable, revealing the outstanding performance of wide-angle FSAR. 
\section{Conclusion}
In this work, we propose Otter which is specially designed against background distractions of wide-angle FSAR. Otter highlights subjects in each frames via adaptive segmentation and enhancement of CSM. Temporal relation degradation caused by too many frames with similar background is reconstructed by bidirectional scanning of TRM. Otter achieves new SOTA performance on several widely-used datasets. Further studies demonstrate the competitiveness of our proposed method, especially for mitigating background distractions of wide-angle FSAR. We hope this work will inspire upcoming research in FSAR community. 

\appendix

\setcounter{figure}{0}
\setcounter{table}{0}
\setcounter{equation}{0}
\renewcommand\thesection{\Alph{section}}
\renewcommand{\thefigure}{\Roman{figure}}
\renewcommand{\thetable}{\Roman{table}}
\renewcommand\theequation{\Roman{equation}}

\section*{Supplementary Materials}
In the supplementary material, we provide:
\begin{itemize}
	\item Extra Study of Key Components~(mentioned in \S~\ref{factor})
	\item Additional Wide-Angle Evaluation~(mentioned in \S~\ref{angle})
	\item Robustness Study
	\item Computational Complexity
\end{itemize}

\section{Extra Study of Key Components}
\subsection{Study on RWKV-4 and RWKV-5/6}
\indent Currently, RWKV-4~\cite{peng2023rwkv} and RWKV-5/6~\cite{peng2024eagle} are released official versions. The main discrepancy is additional gate $G$ mechanism in RWKV-5/6 for the control of information flow. In order to compare the performance, we conduct experiments with three key components under various basis. The results are demonstrated in \tablename~\ref{tab: base model}. We find that applying RWKV-5/6 performs better than components based on RWKV-4. Therefore, we select the updated RWKV-5/6 as the basis of our proposed Otter.  
\begin{table}[ht]
	\centering
	\small
	\setlength{\tabcolsep}{4.4pt}
	\begin{tabular}{l|l|l|cc|cc}
		\Xhline{1pt}
		\multicolumn{1}{c|}{\multirow{2}{*}{S-Mix}} & \multicolumn{1}{c|}{\multirow{2}{*}{T-Mix}} & \multicolumn{1}{c|}{\multirow{2}{*}{C-Mix}} & \multicolumn{2}{c|}{SSv2} & \multicolumn{2}{c}{Kinetics} \\ \cline{4-7} 
		\multicolumn{1}{c|}{} & \multicolumn{1}{c|}{} & \multicolumn{1}{c|}{} & \multicolumn{1}{c|}{1-shot} & 5-shot & \multicolumn{1}{c|}{1-shot} & 5-shot \\ \hline
		R-4 & R-4 & R-4 & \multicolumn{1}{c|}{64.0} & 87.5 & \multicolumn{1}{c|}{89.2} & 94.3 \\ \hline
		R-5/6 & R-4 & R-4 & \multicolumn{1}{c|}{64.2} & 87.4 & \multicolumn{1}{c|}{89.5} & 94.5 \\
		R-4 & R-5/6 & R-4 & \multicolumn{1}{c|}{64.1} & 87.6 & \multicolumn{1}{c|}{89.1} & 94.7 \\
		R-4 & R-4 & R-5/6 & \multicolumn{1}{c|}{64.0} & 87.4 & \multicolumn{1}{c|}{89.4} & 94.4 \\ \hline
		R-5/6 & R-5/6 & R-4 & \multicolumn{1}{c|}{64.2} & 87.8 & \multicolumn{1}{c|}{90.0} & 96.1 \\
		R-5/6 & R-4 & R-5/6 & \multicolumn{1}{c|}{64.4} & 88.1 & \multicolumn{1}{c|}{90.1} & 95.7 \\
		R-4 & R-5/6 & R-5/6 & \multicolumn{1}{c|}{64.2} & 87.9 & \multicolumn{1}{c|}{89.7} & 95.5 \\ \hline
		\rowcolor[HTML]{C0C0C0}R-5/6 & R-5/6 & R-5/6 & \multicolumn{1}{c|}{\textbf{64.7}} & \textbf{88.5} & \multicolumn{1}{c|}{\textbf{90.5}} & \textbf{96.4} \\
		\Xhline{1pt}
	\end{tabular}
	\caption{Comparison~($\uparrow$ Acc.~\%) between RWKV-4~(R-4) and RWKV-5/6~(R-5/6).}
	\label{tab: base model}
\end{table} 
\subsection{Study on Learnable Weights}
\indent In our design of CSM and TRM, learnable weights serve as significant roles in highlighting subjects from background and reconstructing disappearing temporal relation. From the results revealed in \tablename~\ref{tab: weight}, we observe that $lw^{\vartriangle}$ and $\mathring{lw}^{\vartriangle}$ can both improve the performance of wide-angle FSAR. The absence of $lw^{\vartriangle}$ harms the adaptive subjects highlighting while the deficiency of $\mathring{lw}^{\vartriangle}$ damages the bidirectional scanning. Therefore, we devise CSM and TRM both equipped with learnable weights.
\begin{table}[ht]
	\centering
	\small
	\begin{tabular}{c|c|cc|cc}
		\Xhline{1pt}
		\multirow{2}{*}{$lw^{\vartriangle}$} & \multirow{2}{*}{$\mathring{lw}^{\vartriangle}$} & \multicolumn{2}{c|}{SSv2} & \multicolumn{2}{c}{Kinetics} \\ \cline{3-6} 
		&  & \multicolumn{1}{c|}{1-shot} & 5-shot & \multicolumn{1}{c|}{1-shot} & 5-shot \\ \hline
		\ding{55} & \ding{55} & \multicolumn{1}{c|}{61.8} & 85.2 & \multicolumn{1}{c|}{85.7} & 91.1 \\ \hline
		\ding{51} & \ding{55} & \multicolumn{1}{c|}{63.8} & 87.9 & \multicolumn{1}{c|}{89.7} & 95.8 \\
		\ding{55} & \ding{51} & \multicolumn{1}{c|}{62.1} & 86.6 & \multicolumn{1}{c|}{89.4} & 95.1 \\ \hline
		\rowcolor[HTML]{C0C0C0}\ding{51} & \ding{51} & \multicolumn{1}{c|}{\textbf{64.7}} & \textbf{88.5} & \multicolumn{1}{c|}{\textbf{90.5}} & \textbf{96.4} \\
		\Xhline{1pt}
	\end{tabular}
	\caption{Comparison~($\uparrow$ Acc.~\%) of learnable weights.}
	\label{tab: weight}
\end{table}
\subsubsection{Loss Design}
\indent In the loss design, we fix $\mathcal{L}_\text{ce}$ as the primary loss for classification and reveal experiments in \tablename~\ref{tab: design}. As auxiliary loss, both $\mathcal{L}^{1}_{\text{P}}$ and $\mathcal{L}^{1}_{\text{P}}$ combined with $\mathcal{L}_\text{ce}$ can improve the performance via further distinguishing similar classes of prototype. The simultaneous use of the three losses can obtain the best performance of wide-angle FSAR. Therefore, $\mathcal{L}_\text{ce}$, $\mathcal{L}^{1}_{\text{P}}$, and $\mathcal{L}^{1}_{\text{P}}$ are necessary in Otter.
\begin{table}[ht]
	\centering
	\small
	\begin{tabular}{c|c|c|cc|cc}
		\Xhline{1pt}
		\multicolumn{1}{l|}{\multirow{2}{*}{$\mathcal{L} _{\text{ce}}$}} & \multirow{2}{*}{$\mathcal{L}^{1}_{\text{P}}$} & \multirow{2}{*}{$\mathcal{L}^{2}_{\text{P}}$} & \multicolumn{2}{c|}{SSv2} & \multicolumn{2}{c}{Kinetics} \\ \cline{4-7} 
		\multicolumn{1}{l|}{} &  &  & \multicolumn{1}{c|}{1-shot} & 5-shot & \multicolumn{1}{c|}{1-shot} & 5-shot \\ \hline
		\multirow{3}{*}{\ding{51}} & \ding{51} & \ding{55} & \multicolumn{1}{c|}{63.3} & 84.8 & \multicolumn{1}{c|}{89.8} & 95.5 \\
		& \ding{55} & \ding{51} & \multicolumn{1}{c|}{63.4} & 88.0 & \multicolumn{1}{c|}{90.1} & 95.7 \\ \cline{2-7} 
		& \ding{51} & \ding{51} & \multicolumn{1}{c|}{\cellcolor[HTML]{C0C0C0}\textbf{64.7}} & \cellcolor[HTML]{C0C0C0}\textbf{88.5} & \multicolumn{1}{c|}{\cellcolor[HTML]{C0C0C0}\textbf{90.5}} & \cellcolor[HTML]{C0C0C0}\textbf{96.4} \\
		\Xhline{1pt}
	\end{tabular}
	\caption{Comparison~($\uparrow$ Acc.~\%) of loss design.}
	\label{tab: design}
\end{table}
\subsection{Study on Loss Weight Factors}
\indent The training objective is the combination of $\mathcal{L} _{\text{ce}}$, $\mathcal{L}^{1}_{\text{P}}$, and $\mathcal{L}^{2}_{\text{P}}$ with loss weight factors $\lambda$. Experiments are conducted and the results are illustrated in \tablename~\ref{tab: loss}. As a role primarily used for classification, $\lambda_0$ for $\mathcal{L}_\text{ce}$ should not be less than 0.5. Considering the similar function of $\mathcal{L}^{1}_{\text{P}}$ and $\mathcal{L}^{2}_{\text{P}}$, $\lambda_1$ and $\lambda_2$ should be equal. The performance is improved with the increasing $\lambda_0$ but begins to decline when $\lambda_0>0.8$. The above results confirm the loss weight factors.
\begin{table}[ht]
	\centering
	\small
	\begin{tabular}{l|l|l|cc|cc}
		\Xhline{1pt}
		\multicolumn{1}{c|}{\multirow{2}{*}{$\lambda_0$}} & \multicolumn{1}{c|}{\multirow{2}{*}{$\lambda_1$}} & \multicolumn{1}{c|}{\multirow{2}{*}{$\lambda_2$}} & \multicolumn{2}{c|}{SSv2} & \multicolumn{2}{c}{Kinetics} \\ \cline{4-7} 
		&  &  & \multicolumn{1}{c|}{1-shot} & 5-shot & \multicolumn{1}{c|}{1-shot} & 5-shot \\ \hline
		0.50 & 0.25 & 0.25 & \multicolumn{1}{c|}{62.9} & 87.6 & \multicolumn{1}{c|}{89.6} & 95.6 \\
		0.60 & 0.20 & 0.20 & \multicolumn{1}{c|}{64.1} & 88.0 & \multicolumn{1}{c|}{89.9} & 95.9 \\
		0.70 & 0.15 & 0.15 & \multicolumn{1}{c|}{64.3} & 88.2 & \multicolumn{1}{c|}{90.3} & 96.2 \\
		\rowcolor[HTML]{C0C0C0}0.80 & 0.10 & 0.10 & \multicolumn{1}{c|}{\textbf{64.7}} & \textbf{88.5} & \multicolumn{1}{c|}{\textbf{90.5}} & \textbf{96.4} \\
		0.90 & 0.05 & 0.05 & \multicolumn{1}{c|}{64.4} & 88.4 & \multicolumn{1}{c|}{90.2} & 96.2 \\
		\Xhline{1pt}
	\end{tabular}
	\caption{Comparison~($\uparrow$ Acc.~\%) of loss factors.}
	\label{tab: loss}
\end{table}
\subsection{Study on Various Types of Prototype}
\indent There are three types of prototype construction including attention-based calculation~$\mathrm{Attn}\left(\cdot\right)$~\cite{wang2022hybrid}, query-specific prototype~$\mathrm{Q}\text{-}\mathrm{Sp}\left(\cdot\right)$~\cite{perrett2021temporal}, and averaging calculation~$\mathrm{Avg}\left(\cdot\right)$~\cite{huang2025manta}. Experiments about compatibility of Otter and prototype construction is conducted in \tablename~\ref{tab: proto}. Although $\mathrm{Attn}\left(\cdot\right)$ and $\mathrm{Q}\text{-}\mathrm{Sp}\left(\cdot\right)$ with extra calculation achieve advanced performance in their work, the fitness with our Otter is not the best. Therefore, we select simple $\mathrm{Avg}\left(\cdot\right)$ as our prototype.
\begin{table}[ht]
	\centering
	\small
	\begin{tabular}{l|cc|cc}
		\Xhline{1pt}
		\multirow{2}{*}{Prototype} & \multicolumn{2}{c|}{SSv2} & \multicolumn{2}{c}{Kinetics} \\ \cline{2-5} 
		& \multicolumn{1}{c|}{1-shot} & 5-shot & \multicolumn{1}{c|}{1-shot} & 5-shot \\ \hline
		$\mathrm{Attn}\left(\cdot\right)$ & \multicolumn{1}{c|}{63.9} & 87.1 & \multicolumn{1}{c|}{89.3} & 94.9 \\
		$\mathrm{Q}\text{-}\mathrm{Sp}\left(\cdot\right)$ & \multicolumn{1}{c|}{64.5} & 88.5 & \multicolumn{1}{c|}{90.1} & 96.0 \\
		\rowcolor[HTML]{C0C0C0}$\mathrm{Avg}\left(\cdot\right)$ & \multicolumn{1}{c|}{\textbf{64.7}} & \textbf{88.5} & \multicolumn{1}{c|}{\textbf{90.5}} & \textbf{96.4} \\
		\Xhline{1pt}
	\end{tabular}
	\caption{Comparison~($\uparrow$ Acc.~\%) of various prototype types.}
	\label{tab: proto}
\end{table}
\section{Additional Wide-Angle Evaluation}
\label{add}
\subsection{Details of Wider FoV Simulation}
\begin{figure}[ht]
	\centering
	\subcaptionbox{Lv.0\label{fig: r0}}{\includegraphics[width=0.08\textwidth]{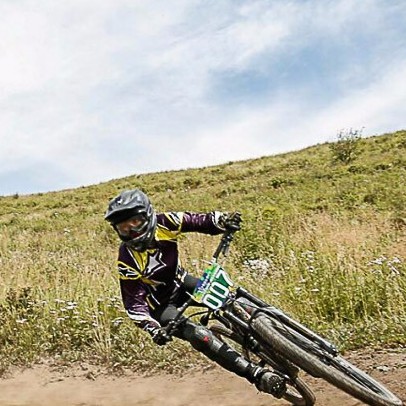}}
	\subcaptionbox{Lv.1\label{fig: r25}}{\includegraphics[width=0.08\textwidth]{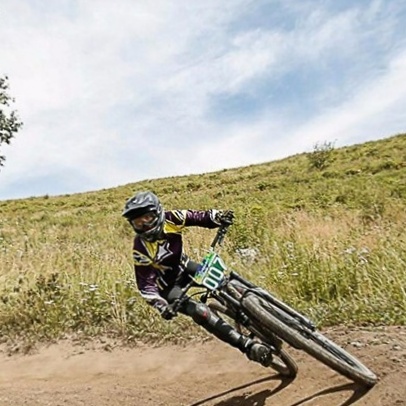}}
	\subcaptionbox{Lv.2\label{fig: r50}}{\includegraphics[width=0.08\textwidth]{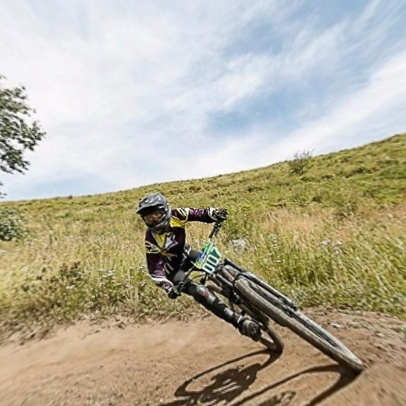}}
	\subcaptionbox{Lv.3\label{fig: r75}}{\includegraphics[width=0.08\textwidth]{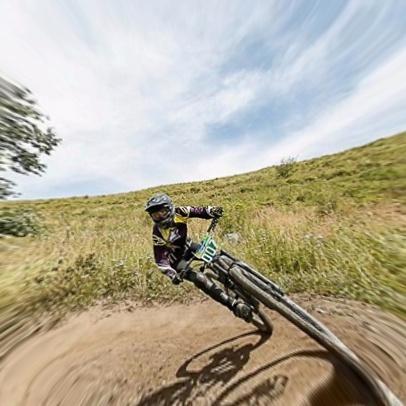}}
	\subcaptionbox{Lv.4\label{fig: r100}}{\includegraphics[width=0.08\textwidth]{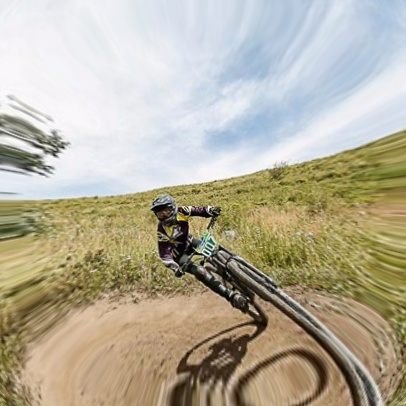}}
	\caption{Examples with various wide FoV levels. To be specific, each level is a combination of $U_\text{m}$ and $U_\text{d}$.}
	\label{fig: fov}
\end{figure}

\indent From previous definition~\cite{liao2023deep}, FoV is only determined by  camera CMOS size~($H_{\text{c}}\times W_{\text{c}}$) and lens focal length~($L_{\text{f}}$). Related calculation is written as
\begin{equation}
	\begin{aligned}
		FoV=2\mathrm{arctan} \left( \frac{\vartriangle}{2L_{\text{f}}} \right) ,\forall \vartriangle \in \left( H_{\text{c}}, W_{\text{c}} \right).
	\end{aligned}
	\label{eq1s}
\end{equation}
Image size is positively correlated with the CMOS size, while the distortion is negatively correlated with the focal length~\cite{hu2022miniature}. Therefore, directly applying larger outpainting magnification~($U_\text{m}$) and introducing larger distortion factor~($U_\text{d}$) can simulate wider FoV. A group of simulation with five various levels is provided in \figurename~\ref{fig: fov}. We observe that a wider FoV means more background. Meanwhile, distortion is more exaggerated. Wide-angle datasets always correct them for stable training. Re-adding distortion makes wide-angle FSAR more challenging.
\subsection{Temporal Relation}
\begin{figure}[ht]
	\centering
	\subcaptionbox{1-shot\label{fig: dtw_ssv2}}{\includegraphics[width=0.23\textwidth]{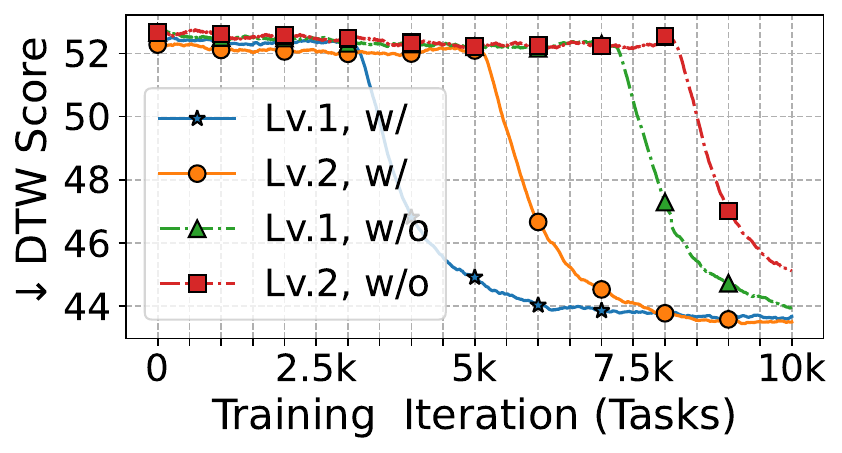}}
	\subcaptionbox{5-shot\label{fig: dtw_kinetics}}{\includegraphics[width=0.23\textwidth]{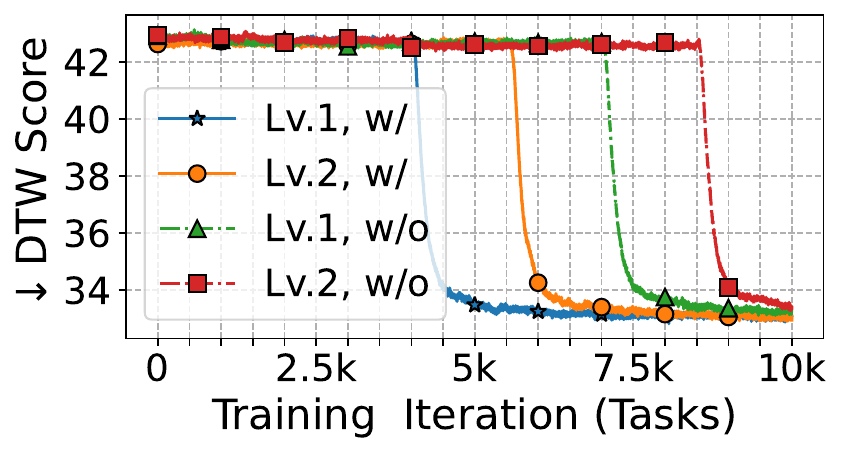}}
	\caption{$\downarrow$ DTW scores under 1-shot setting during training on VideoBadminton. In specific, “Lv.1, w/” denotes model equipped with Otter and samples with Lv.1 FoV.}
	\label{fig: dtw}
\end{figure}

\indent According to OTAM~\cite{cao2020few}, DTW scores calculated from two sequences~($\downarrow$ lower indicates better) can reflect the quality of temporal relation via alignment degree. The curves are shown in \figurename~\ref{fig: dtw}. We observe that models equipped with Otter converge much faster than those without Otter under any few-shot setting. The convergence points for the 5-shot are much earlier due to the increased number of training samples. Under the 1-shot setting of FSAR, the DTW curves without Otter even do not converge under Lv.1 or 2 FoV, indicating a more time-consuming training. Therefore, it is evident that Otter can effectively reconstruct temporal relations of wide-angle FSAR.
\subsection{T-SNE Visualization}
\begin{figure}[ht]
	\centering
	\subcaptionbox{w/o Otter\label{fig: without}}{\includegraphics[width=0.16\textwidth]{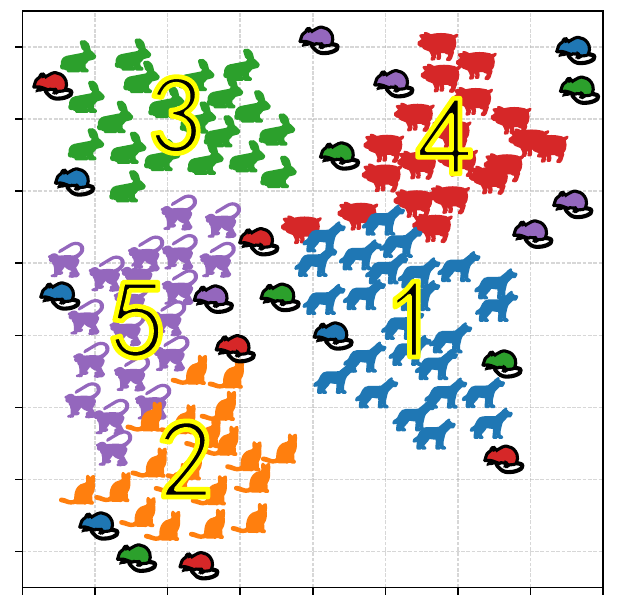}}
	\subcaptionbox{w/ Otter\label{fig: with}}{\includegraphics[width=0.16\textwidth]{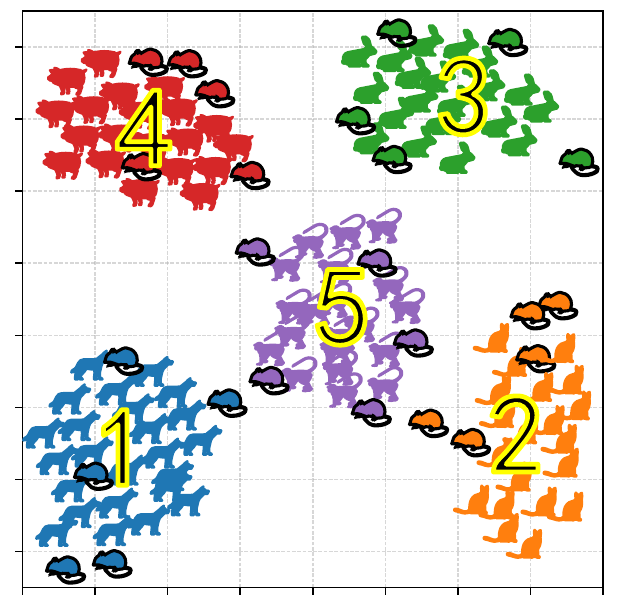}}
	\caption{T-SNE visualization of five action classes in support from Kinetics~(25-shot). \textcolor[HTML]{1F77B4}{Blue}: “ice skating”, \textcolor[HTML]{FF7F0E}{Orange}: “snowboarding”, \textcolor[HTML]{2CA02C}{Green}: “paragliding”, \textcolor[HTML]{D62728}{Red}: “skateboarding”, \textcolor[HTML]{9467BD}{Purple}: “crossing river”. Dots with black borders are samples with Lv.4 FoV simulation.}
	\label{fig: tsne}
\end{figure}

\indent From the t-SNE~\cite{van2008visualizing} revealed in \figurename~\ref{fig: tsne}, the wide-angle actions are hard to be separated and clustered well without any assistance. Samples with Lv.4 FoV simulation are scattered everywhere. The above observation prove the difficulties in wide-angle FSAR. On the contrary, Otter clusters samples from same class and scatters others better. Although these special samples with 100\% expanding magnification are located at the edge of each class, the cluster condition of them is much better.
\subsection{Additional CAM Visualization}
\begin{figure}[t]
	\centering
	\subcaptionbox{crossing river\label{fig: crossing_river}}{\includegraphics[width=0.42\textwidth]{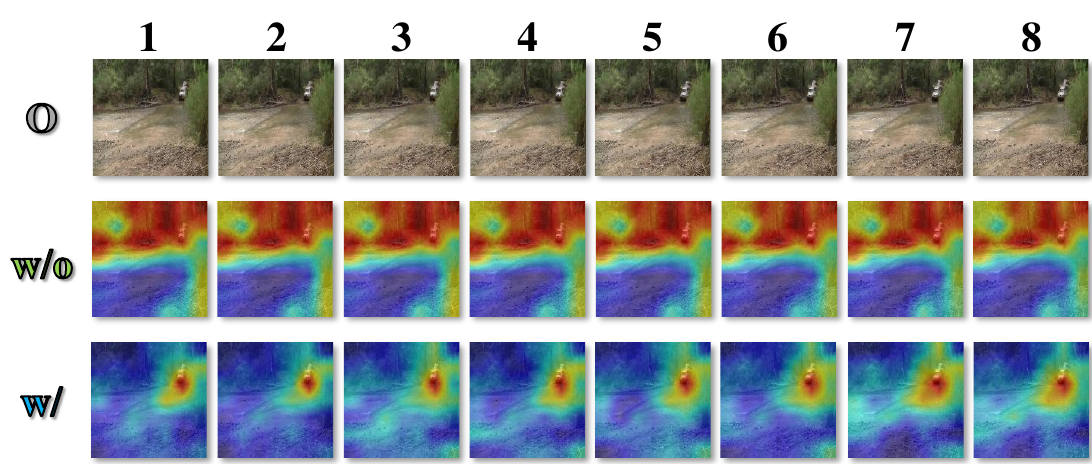}}
	\subcaptionbox{paragliding\label{fig: paragliding}}{\includegraphics[width=0.42\textwidth]{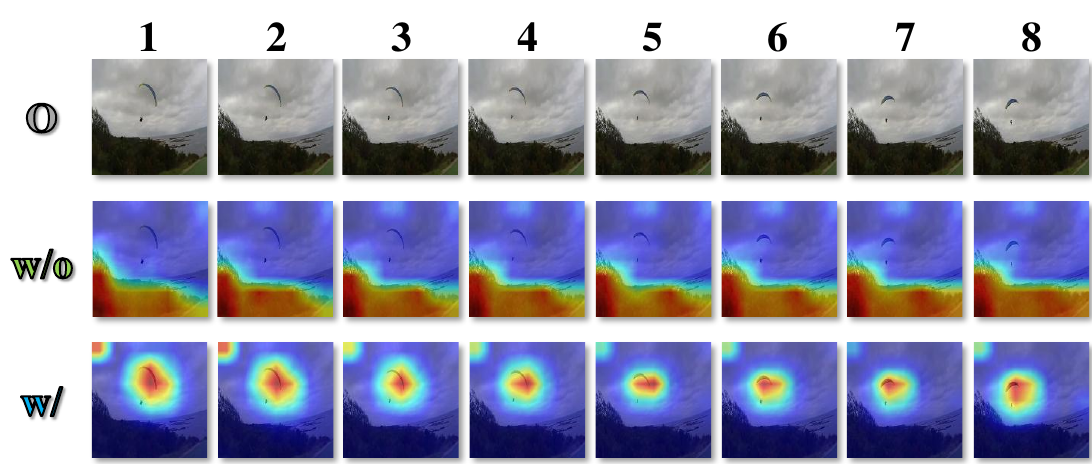}}
	\subcaptionbox{driving tractor\label{fig: driving_tractor}}{\includegraphics[width=0.42\textwidth]{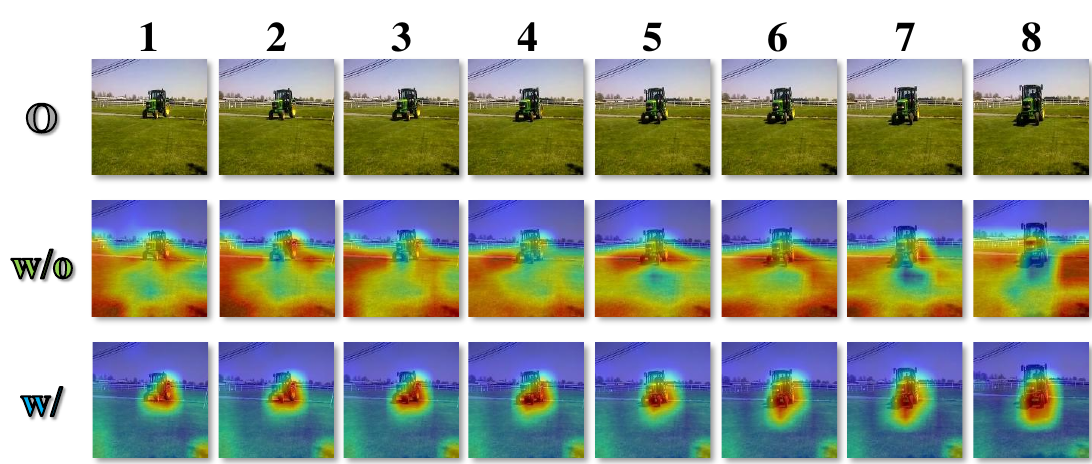}}
	\caption{Additional CAM of “crossing river”, “paragliding”, and “driving tractor” from Kinetics dataset.}
	\label{fig: addition}
\end{figure}

\indent Additional CAM visualization for wide-angle samples are provided in \figurename~\ref{fig: addition}. Taking “crossing river” as an example, it is evident that the model without Otter focuses on “forests” due to their larger proportion in the frames. Although subject “Jeep” is included, recognition is inevitably interfered with by the background. In contrast, Otter accurately highlights the subject while not completely ignoring the background, thereby achieving better performance. This focus pattern is consistent across the other two examples. These CAM visualization demonstrate that Otter mitigates background distractions, helping models better understand challenging actions in wide-angle scenario.
\section{Robustness Study}
\begin{figure}[ht]
	\centering
	\subcaptionbox{O\label{fig: threecat}}{\includegraphics[width=0.08\textwidth]{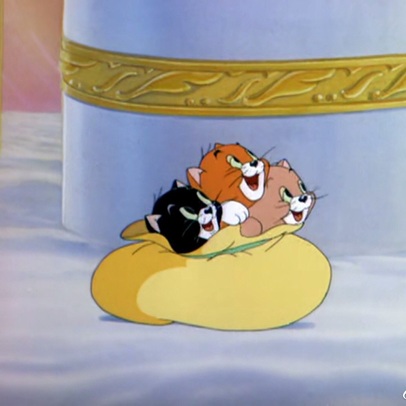}}
	\subcaptionbox{Z\label{fig: 2x}}{\includegraphics[width=0.08\textwidth]{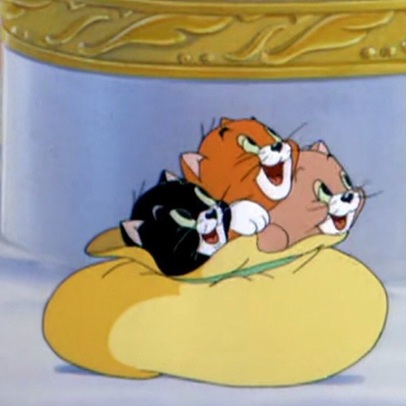}}
	\subcaptionbox{G\label{fig: gaussian}}{\includegraphics[width=0.08\textwidth]{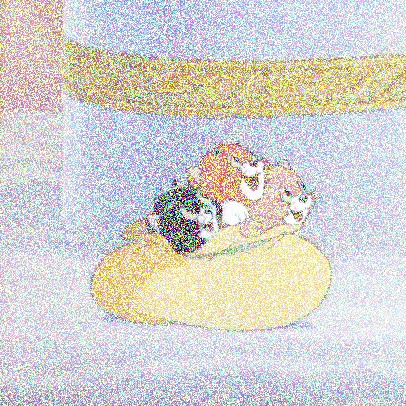}}
	\subcaptionbox{R\label{fig: rainy}}{\includegraphics[width=0.08\textwidth]{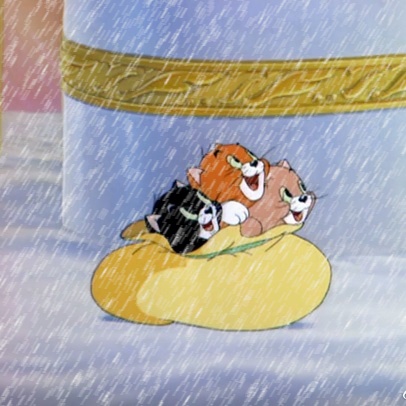}}
	\subcaptionbox{L\label{fig: light}}{\includegraphics[width=0.08\textwidth]{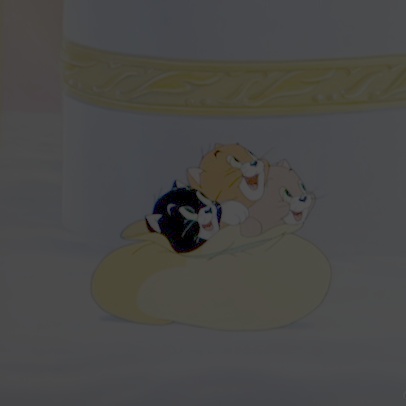}}
	\caption{Different kinds of noise. In specific, O, Z, G, R, and L denote original frames, zoom, Gaussian, rainy, and light noise, respectively.}
	\label{fig: noise}
\end{figure}

\indent In order to explore robustness of Otter, we select two groups of noise added into $\mathcal{D}_\text{test}$ of FSAR. The first group is task-based including sample-level and frame-level noise for simulating unexpected circumstances during data collection. As revealed in \figurename~\ref{fig: noise}, another group is visual noise such as zoom, Gaussian, rainy, and light noise, for simulating different shooting situations. Specifically, zoom frames are imposed by variation in optimal zoom while Gaussian noise is related to digital issues of hardware. Changeable weather and lighting conditions result in rainy and light noise. 
\subsection{Sample-Level Noise}
\indent Wide-angle samples from other classes may be mixed into a particular class. Correcting sample-level noise is time-consuming and laborious. Therefore, directly testing wide-angle FSAR on sample-level noise can reflect the robustness of a method. The experimental results are indicated in \tablename~\ref{tab: sample}. It is obvious that the introduce of sample-level has negative impacts on the performance of wide-angle FSAR. The results decline with the increasing ratio of sample-level noise. However, we find that the robustness of our proposed Otter is better than other recent methods.
\begin{table}[ht]
	\centering
	\small
	\setlength{\tabcolsep}{4.4pt}
	\begin{tabular}{l|l|ccccc}
		\Xhline{1pt}
		\multirow{2}{*}{Datasets} & \multirow{2}{*}{Methods} & \multicolumn{5}{c}{Sample-Level Noise Ratio} \\ \cline{3-7} 
		&  & \multicolumn{1}{c|}{0\%} & \multicolumn{1}{c|}{10\%} & \multicolumn{1}{c|}{20\%} & \multicolumn{1}{c|}{30\%} & 40\% \\ \hline
		\multirow{4}{*}{SSv2} & MoLo & \multicolumn{1}{c|}{72.5} & \multicolumn{1}{c|}{70.5} & \multicolumn{1}{c|}{68.2} & \multicolumn{1}{c|}{66.4} & 64.1 \\
		& SOAP & \multicolumn{1}{c|}{87.3} & \multicolumn{1}{c|}{85.1} & \multicolumn{1}{c|}{83.0} & \multicolumn{1}{c|}{80.8} & 78.7 \\
		& Manta & \multicolumn{1}{c|}{89.6} & \multicolumn{1}{c|}{87.6} & \multicolumn{1}{c|}{86.2} & \multicolumn{1}{c|}{83.1} & 80.9 \\ \cline{2-7} 
		&\cellcolor[HTML]{C0C0C0}Otter & \multicolumn{1}{c|}{\cellcolor[HTML]{C0C0C0}\textbf{90.2}} & \multicolumn{1}{c|}{\cellcolor[HTML]{C0C0C0}\textbf{89.4}} & \multicolumn{1}{c|}{\cellcolor[HTML]{C0C0C0}\textbf{88.2}} & \multicolumn{1}{c|}{\cellcolor[HTML]{C0C0C0}\textbf{86.6}} & \cellcolor[HTML]{C0C0C0}\textbf{85.5} \\ \hline
		\multirow{4}{*}{Kinetics} & MoLo & \multicolumn{1}{c|}{87.5} & \multicolumn{1}{c|}{85.1} & \multicolumn{1}{c|}{83.4} & \multicolumn{1}{c|}{80.8} & 78.1 \\
		& SOAP & \multicolumn{1}{c|}{95.9} & \multicolumn{1}{c|}{94.2} & \multicolumn{1}{c|}{92.1} & \multicolumn{1}{c|}{89.7} & 87.5 \\
		& Manta & \multicolumn{1}{c|}{96.1} & \multicolumn{1}{c|}{94.2} & \multicolumn{1}{c|}{91.9} & \multicolumn{1}{c|}{90.1} & 87.8 \\ \cline{2-7} 
		& \cellcolor[HTML]{C0C0C0}Otter & \multicolumn{1}{c|}{\cellcolor[HTML]{C0C0C0}\textbf{98.4}} & \multicolumn{1}{c|}{\cellcolor[HTML]{C0C0C0}\textbf{97.5}} & \multicolumn{1}{c|}{\cellcolor[HTML]{C0C0C0}\textbf{96.2}} & \multicolumn{1}{c|}{\cellcolor[HTML]{C0C0C0}\textbf{95.0}} & \cellcolor[HTML]{C0C0C0}\textbf{93.8} \\
		\Xhline{1pt}
	\end{tabular}
	\caption{Comparison~($\uparrow$ Acc.~\%) with sample-level noise under 5-way 10-shot setting.}
	\label{tab: sample}
\end{table}
\subsection{Frame-Level Noise}
\indent Multiple irrelevant frames mixed into wide-angle samples are called as frame-level noise. Serving as a unexpected situation of data collection, robustness of methods can also be reflected by frame-level noise. From the results in \tablename~\ref{tab: frame}, we observe that the performance of wide-angle FSAR is harmed with the increasing number of noisy frames. The reason for this phenomenon is that frame-level noise further disorganizes subjects and temporal relation. Under the circumstance, our Otter still reveals stable performance, reflecting better robustness of frame-level noise.   
\begin{table}[ht]
	\centering
	\small
	\setlength{\tabcolsep}{4.4pt}
	\begin{tabular}{l|l|ccccc}
		\Xhline{1pt}
		\multirow{2}{*}{Datasets} & \multirow{2}{*}{Methods} & \multicolumn{5}{c}{Noisy Frame Numbers} \\ \cline{3-7} 
		&  & \multicolumn{1}{c|}{0} & \multicolumn{1}{c|}{1} & \multicolumn{1}{c|}{2} & \multicolumn{1}{c|}{3} & 4 \\ \hline
		\multirow{4}{*}{SSv2} & MoLo & \multicolumn{1}{c|}{72.5} & \multicolumn{1}{c|}{69.3} & \multicolumn{1}{c|}{66.5} & \multicolumn{1}{c|}{63.3} & 59.6 \\
		& SOAP & \multicolumn{1}{c|}{87.3} & \multicolumn{1}{c|}{84.1} & \multicolumn{1}{c|}{80.9} & \multicolumn{1}{c|}{78.0} & 75.6 \\
		& Manta & \multicolumn{1}{c|}{89.6} & \multicolumn{1}{c|}{86.4} & \multicolumn{1}{c|}{83.2} & \multicolumn{1}{c|}{80.4} & 77.3 \\ \cline{2-7} 
		& \cellcolor[HTML]{C0C0C0}Otter & \multicolumn{1}{c|}{\cellcolor[HTML]{C0C0C0}\textbf{90.2}} & \multicolumn{1}{c|}{\cellcolor[HTML]{C0C0C0}\textbf{89.0}} & \multicolumn{1}{c|}{\cellcolor[HTML]{C0C0C0}\textbf{88.2}} & \multicolumn{1}{c|}{\cellcolor[HTML]{C0C0C0}\textbf{87.2}} & \cellcolor[HTML]{C0C0C0}\textbf{86.0} \\ \hline
		\multirow{4}{*}{Kinetics} & MoLo & \multicolumn{1}{c|}{87.5} & \multicolumn{1}{c|}{84.3} & \multicolumn{1}{c|}{81.5} & \multicolumn{1}{c|}{78.0} & 75.3 \\
		& SOAP & \multicolumn{1}{c|}{95.9} & \multicolumn{1}{c|}{93.1} & \multicolumn{1}{c|}{90.2} & \multicolumn{1}{c|}{87.6} & 84.1 \\
		& Manta & \multicolumn{1}{c|}{96.1} & \multicolumn{1}{c|}{93.0} & \multicolumn{1}{c|}{89.7} & \multicolumn{1}{c|}{86.8} & 83.7 \\ \cline{2-7} 
		& \cellcolor[HTML]{C0C0C0}Otter & \multicolumn{1}{c|}{\cellcolor[HTML]{C0C0C0}\textbf{98.4}} & \multicolumn{1}{c|}{\cellcolor[HTML]{C0C0C0}\textbf{97.1}} & \multicolumn{1}{c|}{\cellcolor[HTML]{C0C0C0}\textbf{95.7}} & \multicolumn{1}{c|}{\cellcolor[HTML]{C0C0C0}\textbf{94.5}} & \cellcolor[HTML]{C0C0C0}\textbf{93.2} \\
		\Xhline{1pt}
	\end{tabular}
	\caption{Comparison~($\uparrow$ Acc.~\%) with frame-level noise under 5-way 10-shot setting.}
	\label{tab: frame}
\end{table}
\subsection{Visual-Based Noise}
\indent Visual-based noise challenges the robustness of a method. Therefore, we add each type of visual-based noise to 25\% samples for creating more complex wide-angle FSAR tasks. As shown in \tablename~\ref{tab: visual}, the zoom noise has the largest negative impact on the performance. Other types of visual-based noise more or less harm the results. However, we observe that our Otter can keep the SOTA performance under those challenging environment. These phenomena in wide-angle FSAR reflect the better robustness of the proposed Otter. 
\begin{table}[ht]
	\centering
	\small
	\setlength{\tabcolsep}{4.4pt}
	\begin{tabular}{l|l|ccccc}
		\Xhline{1pt}
		\multirow{2}{*}{Datasets} & \multirow{2}{*}{Methods} & \multicolumn{5}{c}{Visual-Based Noise Type} \\ \cline{3-7} 
		&  & \multicolumn{1}{c|}{O} & \multicolumn{1}{c|}{Z} & \multicolumn{1}{c|}{G} & \multicolumn{1}{c|}{R} & L \\ \hline
		\multirow{4}{*}{SSv2} & MoLo & \multicolumn{1}{c|}{72.5} & \multicolumn{1}{c|}{70.0} & \multicolumn{1}{c|}{70.3} & \multicolumn{1}{c|}{69.7} & 69.8 \\
		& SOAP & \multicolumn{1}{c|}{87.3} & \multicolumn{1}{c|}{84.7} & \multicolumn{1}{c|}{84.0} & \multicolumn{1}{c|}{84.6} & 86.1 \\
		& Manta & \multicolumn{1}{c|}{89.6} & \multicolumn{1}{c|}{87.5} & \multicolumn{1}{c|}{88.7} & \multicolumn{1}{c|}{88.8} & 87.4 \\ \cline{2-7} 
		& \cellcolor[HTML]{C0C0C0}Otter & \multicolumn{1}{c|}{\cellcolor[HTML]{C0C0C0}\textbf{90.2}} & \multicolumn{1}{c|}{\cellcolor[HTML]{C0C0C0}\textbf{89.6}} & \multicolumn{1}{c|}{\cellcolor[HTML]{C0C0C0}\textbf{89.6}} & \multicolumn{1}{c|}{\cellcolor[HTML]{C0C0C0}\textbf{89.3}} & \cellcolor[HTML]{C0C0C0}\textbf{89.0} \\ \hline
		\multirow{4}{*}{Kinetics} & MoLo & \multicolumn{1}{c|}{87.5} & \multicolumn{1}{c|}{85.2} & \multicolumn{1}{c|}{86.3} & \multicolumn{1}{c|}{86.7} & 85.9 \\
		& SOAP & \multicolumn{1}{c|}{95.9} & \multicolumn{1}{c|}{93.6} & \multicolumn{1}{c|}{94} & \multicolumn{1}{c|}{94.4} & 93.9 \\
		& Manta & \multicolumn{1}{c|}{96.1} & \multicolumn{1}{c|}{93.9} & \multicolumn{1}{c|}{95.0} & \multicolumn{1}{c|}{95.1} & 94.8 \\ \cline{2-7} 
		& \cellcolor[HTML]{C0C0C0}Otter & \multicolumn{1}{c|}{\cellcolor[HTML]{C0C0C0}\textbf{98.4}} & \multicolumn{1}{c|}{\cellcolor[HTML]{C0C0C0}\textbf{97.9}} & \multicolumn{1}{c|}{\cellcolor[HTML]{C0C0C0}\textbf{98.0}} & \multicolumn{1}{c|}{\cellcolor[HTML]{C0C0C0}\textbf{97.7}} & \cellcolor[HTML]{C0C0C0}\textbf{97.8} \\
		\Xhline{1pt}
	\end{tabular}
	\caption{Comparison~($\uparrow$ Acc.~\%) with various types of 25\% visual-based noise under 5-way 10-shot setting.}
	\label{tab: visual}
\end{table}
\subsection{Cross Dataset Testing}
\indent In real-world scenario, various data distributions are exist. Therefore, we applying the cross dataset method~(training and testing on various datasets) for the simulation of different data distributions. SSv2 and Kinetics with three no-overlapping set are utilized. Then overlapping classes of $\mathcal{D}_\text{train}$ and $\mathcal{D}_\text{test}$ from different datasets are further removed. From the results revealed in \tablename~\ref{tab: cross}, despite cross-dataset setting degrades the performance, Otter can keep ahead of other methods. This trend similar with the regular test setting highlights the robustness of Otter.
\begin{table}[ht]
	\centering
	\small
	\setlength{\tabcolsep}{4.4pt}
	\begin{tabular}{l|cc|cc}
		\Xhline{1pt}
		\multirow{2}{*}{Methods} & \multicolumn{2}{c|}{KI$\rightarrow$SS \scriptsize(SS$\rightarrow$SS)} & \multicolumn{2}{c}{SS$\rightarrow$KI \scriptsize(KI$\rightarrow$KI)} \\ \cline{2-5} 
		& \multicolumn{1}{c|}{1-shot} & 5-shot & \multicolumn{1}{c|}{1-shot} & 5-shot \\ \hline
		MoLo & \multicolumn{1}{c|}{53.7 \scriptsize(56.6)} & 68.7 \scriptsize(70.7) & \multicolumn{1}{c|}{71.5 \scriptsize(74.2)} & 83.2 \scriptsize(85.7) \\
		SOAP & \multicolumn{1}{c|}{60.0 \scriptsize(61.9)} & 84.5 \scriptsize(85.8) & \multicolumn{1}{c|}{84.1 \scriptsize(86.1)} & 91.1 \scriptsize(93.8) \\
		Manta & \multicolumn{1}{c|}{61.5 \scriptsize(63.4)} & 86.4 \scriptsize(87.4) & \multicolumn{1}{c|}{86.3 \scriptsize(87.4)} & 91.8 \scriptsize(94.2) \\ \hline
		\rowcolor[HTML]{C0C0C0}Otter & \multicolumn{1}{c|}{\textbf{63.1} \scriptsize\textbf{(64.7)}} & \textbf{86.7} \scriptsize\textbf{(88.5)} & \multicolumn{1}{c|}{\textbf{89.2} \scriptsize\textbf{(90.5)}} & \textbf{94.0 \scriptsize\textbf{(96.4)}} \\ 
		\Xhline{1pt}
	\end{tabular}
	\caption{Comparison~($\uparrow$ Acc.~\%) with cross dataset~(large fonts) and regular testing~(small fonts in brackets). KI$\rightarrow$SS: Kinetics training while SSv2 testing, SS$\rightarrow$KI: SSv2 training while Kinetics testing, SS$\rightarrow$SS: training and testing both on SSv2, KI$\rightarrow$KI: training and testing both on Kinetics.}
	\label{tab: cross}
\end{table}
\subsection{Any-Shot Testing}
\indent In real-world application, ensuring shot number of each class equal is challenging. In order to create a more authentic testing environment for robustness, we apply the any-shot setup~($1\leqslant K\leqslant 5$). From the results demonstrated in \tablename~\ref{tab: any}, we observe that the performance of Otter defeats other methods, reflecting our Otter has a better robustness for applications in real-world scenario.
\begin{table}[ht]
	\centering
	\small
	\begin{tabular}{l|c|c}
		\Xhline{1pt}
		Methods & SSv2 & Kinetics \\ \hline
		MoLo & 64.6~\scriptsize{$\pm \text{1.5}$}& 80.2~\scriptsize{$\pm \text{1.8}$} \\
		SOAP & 73.8~\scriptsize{$\pm \text{1.4}$} & 89.1~\scriptsize{$\pm \text{1.3}$} \\
		Manta & 75.2~\scriptsize{$\pm \text{1.1}$} & 90.6~\scriptsize{$\pm \text{1.3}$} \\ \hline
		\rowcolor[HTML]{C0C0C0}Otter & \textbf{77.4~\scriptsize{$\pm \text{0.6}$}} & \textbf{93.6~\scriptsize{$\pm \text{0.7}$}} \\
		\Xhline{1pt}
	\end{tabular}
	\caption{Comparison~($\uparrow$ Acc.~\%) with 95\% confidence interval of 5-way any-shot setting.}
	\label{tab: any}
\end{table}
\section{Computational Complexity}
\subsection{Inference Speed}
\indent To evaluate the model under practical conditions with limited resources, we conducted 10,000 tasks using a single 24GB NVIDIA GeForce RTX 3090 GPU on a server. From results demonstrated in \tablename~\ref{tab: single}, we find that the inference speed of MoLo and SOAP is slow because of Transformer with high computational complexity. On the contrary, Mamba-based Manta and RWKV-based Otter is much faster than previous Transformer-based methods. Considering the accuracy of classification, the proposed Otter is more suitable for practical applications.
\begin{table}[ht]
	\centering
	\small
	\begin{tabular}{l|cc|cc}
		\Xhline{1pt}
		\multirow{2}{*}{Methods} & \multicolumn{2}{c|}{SSv2} & \multicolumn{2}{c}{Kinetics} \\ \cline{2-5} 
		& \multicolumn{1}{c|}{1-shot} & 5-shot & \multicolumn{1}{c|}{1-shot} & 5-shot \\ \hline
		MoLo & \multicolumn{1}{c|}{7.83} & 8.02 & \multicolumn{1}{c|}{7.64} & 8.14 \\
		SOAP & \multicolumn{1}{c|}{7.44} & 7.86 & \multicolumn{1}{c|}{7.21} & 7.72 \\
		Manta & \multicolumn{1}{c|}{4.25} & 4.61 & \multicolumn{1}{c|}{4.42} & 4.56 \\ \hline
		\rowcolor[HTML]{C0C0C0}Otter & \multicolumn{1}{c|}{\textbf{4.13}} & \textbf{4.24} & \multicolumn{1}{c|}{\textbf{4.35}} & \textbf{4.48} \\ 
		\Xhline{1pt}
	\end{tabular}
	\caption{Inference speed~($\downarrow$~hour) with 10,000 random tasks on single 24GB NVIDIA GeForce RTX 3090 GPU.}
	\label{tab: single}
\end{table} 
\subsection{Major Tensor Changes}
\indent The tensor changes detailed in \tablename~\ref{tab: structure} offer deeper insights into Otter. For simplicity, we use the wildcard symbol $\vartriangle$ as in the main paper. These tensor changes facilitate the determination of hyper-parameters, such as the patch size~($p=56$). Additionally, we observe that the primary computational burden lies in the $\mathrm{Seg}\left( \cdot, \cdot \right)$ and $\mathrm{RT}\left( \dots ,\cdot ,\dots \right)$ components of the CSM, confirming the single-scale patch design for reducing computational cost. In the following pseudo-code, we provide the further analysis of computational complexity in the proposed Otter.
\begin{table}[ht]
	\centering
	\small
	\setlength{\tabcolsep}{0.8pt}
	\begin{tabular}{l|c|r|c|r}
		\Xhline{1pt}
		Operation & Input & \multicolumn{1}{c|}{Input Size} & Output & \multicolumn{1}{c}{Output Size} \\ \hline
		$\mathrm{Seg}$ & $\vartriangle$ & [$F$, $C$, $H$, $W$] & $\vartriangle ^p$ & [$F$, $C$, $p$, $p$] \\ \hline
		$\mathrm{RT}$ & $lw^{\vartriangle}\odot \vartriangle ^{\beta}$ & [$F$, $C$, $p$, $p$] & $\dot{\vartriangle}$ & [$F$, $C$, $H$, $W$] \\ \hline
		CSM & $S^{c,k},Q^{\gamma}$ & [$F$, $C$, $H$, $W$] & $\hat{S}^{c,k}, \hat{Q}^{\gamma }$ & [$F$, $C$, $H$, $W$] \\ \hline
		$f_{\theta}$ & $\hat{S}^{c,k}, \hat{Q}^{\gamma }$ & [$F$, $C$, $H$, $W$] & $S^{c,k}_{f}, Q^{\gamma}_{f}$ & [$F$, $D$] \\ \hline
		TRM & $\mathring{lw}^{\vartriangle}\odot \mathring{\vartriangle}$ & [$F$, $D$] & $\grave{\vartriangle}$ & [$F$, $D$] \\
		\Xhline{1pt}
	\end{tabular}
	\caption{Major tensor changes in the proposed Otter. Wildcard symbol $\vartriangle$ is applied for simple demonstration and notions are consistent with the main paper.}
	\label{tab: structure}
\end{table} 
\subsection{Pseudo-Code}
\indent The primary computational burden lies in the Compound Segmentation Module (CSM). For complexity analysis, the related pseudo-code is listed in Algorithm~\ref{alg: MaM}. Considering the low computational complexity of core units including $\mathrm{S}\text{-}\mathrm{Mix}\left( \cdot \right)$, $\mathrm{T}\text{-}\mathrm{Mix}\left( \cdot \right)$, and $\mathrm{C}\text{-}\mathrm{Mix}\left( \cdot \right)$ in RWKV, the functions $\mathrm{Seg}(\cdot, \cdot)$ and $\mathrm{RT}(\dots, \cdot, \dots)$ form the main structure with nested loops. Both the inner and outer loops have a computational complexity of $O(p)$. Consequently, the total complexity of the CSM is $O(p^2)$. Given the determined size and single-scale design of $p$~$(p={56})$, the additional computational burden introduced by Otter is negligible, ensuring its usability in real-world applications.
\begin{algorithm}[t]
	\small
	\caption{Compound Motion Segmentation}
	\label{alg: MaM}
	\LinesNumbered
	\SetKwFunction{SMix}{S-Mix}
	\SetKwFunction{CMix}{C-Mix}
	\SetKwFunction{Conv}{Conv}
	\KwIn{$\vartriangle \in \mathbb{R} ^{F\times C\times H\times W}, \forall{\vartriangle} \in \left\{ S^{c,k},Q^{\gamma} \right\}$,  $NP~\left( H\mid NP, W\mid NP, NP\in \mathbb{Z} ^+\right)$.}
	\KwOut{$\hat{\vartriangle} \in \mathbb{R} ^{F\times C\times H\times W}, \forall{\vartriangle} \in \left\{ S^{c,k},Q^{\gamma} \right\}$.}
	$\hat{\vartriangle}_1\gets \varnothing $\;
	\If{$H \% NP ==0$ $\textbf{\&}$ $W \% NP == 0$}
	{$Block \gets \left(H / NP, W / NP \right)$ \;
		\If{$Block\left[0\right] == Block\left[1\right]$}
		{$p_H \gets Block\left[0\right]$ \;
			$p_W \gets Block\left[1\right]$ \;
			\For{$\textbf{each}$ $i \in \left[0, p_H\right]$}
			{\For{$\textbf{each}$ $j \in \left[0, p_W\right]$}
				{\tcc{$\mathrm{Seg}\left( \cdot, \cdot \right)$ $\textbf{Start}$}
					$s \gets \left[i\times p_H, j\times p_W\right]$\;
					$e \gets \left[\left(i+1\right)\times p_H, \left(j+1\right)\times p_W\right]$\;
					$p \gets \vartriangle\left[:, :, s\left(0\right):e\left(0\right), s\left(1\right):e\left(1\right)\right]$\;
					$p1 \gets \SMix\left(p\right) \oplus p$ \;
					$p2 \gets \CMix\left(p1\right) \oplus p1$ \;
					$lw \gets \sigma\left[ \Conv\left(p2\right) \oplus p \right]$ \;
					$\hat{p} \gets lw\odot p2$ \;
					\tcc{$\mathrm{Seg}\left( \cdot, \cdot \right)$ $\textbf{End}$}
					\tcc{$\mathrm{RT}\left( \dots ,\cdot ,\dots \right)$ $\textbf{Start}$}
					$\hat{\vartriangle}_1\left[:,:,s\left(0\right):e\left(0\right), s\left(1\right):e\left(1\right)\right] \gets \hat{p}$\;
					\tcc{$\mathrm{RT}\left( \dots ,\cdot ,\dots \right)$ $\textbf{End}$}
				}
				$\hat{\vartriangle}_2 \gets \SMix\left(\hat{\vartriangle}_1\right) \oplus \hat{\vartriangle}_1$\;
				$\hat{\vartriangle} \gets \CMix\left(\hat{\vartriangle}_2\right) \oplus \hat{\vartriangle}_2$\;
			}
		}
	}
	\Return{$\hat{\vartriangle}$}
\end{algorithm}
\section{Contribution Statement}
\indent This work represents a \textbf{collaborative effort} among all authors, each contributing expertise from different perspectives. The specific contributions are as follows:
\begin{itemize}
	\item \textbf{Wenbo~Huang~(Southeast University, China; Institute of Science Tokyo, Japan):} Firstly proposing the idea of applying RWKV in FSAR, implementing all code of Otter, designing wide-angle evaluation in \S~\ref{add}, conducting all experiments, deriving all mathematical formulas, all data collection, all figure drawing, all table organizing, and completing original manuscript.
	\item \textbf{Jinghui~Zhang~(Southeast University, China):} Providing experimental platform in China, supervision in China, writing polish mainly on logic of introduction, checking results, and funding acquisition.
	\item \textbf{Zhenghao~Chen~(The University of Newcastle, Australia):} Writing polish mainly on authentic expression, clarifying the definition of wide-angle, amending mathematical formulas, checking experimental results, rebuttal assistance, and funding acquisition.
	\item \textbf{Guang~Li~(Hokkaido University, Japan):} Refining Idea of Otter, writing polish mainly on descriptions of results, rebuttal assistance, and funding acquisition.
	\item \textbf{Lei~Zhang~(Nanjing Normal University, China):} Proposing the extra experiments on real wide-angle dataset VideoBadminton, guiding CAM visualization, rebuttal assistance, and funding acquisition.
	\item \textbf{Yang~Cao~(Institute of Science Tokyo, Japan):} Verification of the overall structure, providing experimental platform in Japan, supervision in Japan, rebuttal assistance, and funding acquisition.
	\item \textbf{Fang~Dong~(Southeast University, China):} Funding acquisition.
	\item \textbf{Takahiro~Ogawa~(Hokkaido University, Japan):} Writing polish mainly on numerous details, guiding t-SNE visualization, rebuttal assistance, and funding acquisition.
	\item \textbf{Miki~Haseyama~(Hokkaido University, Japan):} Funding acquisition.
\end{itemize}
\section*{Acknowledgments}
\indent The authors would like to appreciate all participants of peer review and cloud servers provided by Paratera Ltd. Wenbo Huang sincerely thanks those who offered companionship and encouragement during the most challenging times, even though life has since taken everyone on different paths. This work is supported by Frontier Technologies Research and Development Program of Jiangsu under Grant No.~BF2024070; National Natural Science Foundation of China under Grants Nos.~62472094, 62072099, 62232004, 62373194, 62276063; Jiangsu Provincial Key Laboratory of Network and Information Security under Grant No.~BM2003201; Key Laboratory of Computer Network and Information Integration~(Ministry of Education, China) under Grant No.~93K-9; the Fundamental Research Funds for the Central Universities; JSPS KAKENHI Nos.~JP23K21676, JP24K02942, JP24K23849, JP25K21218, JP23K24851; JST PRESTO Grant No.~JPMJPR23P5; JST CREST Grant No.~JPMJCR21M2; JST NEXUS Grant No.~JPMJNX25C4; and Startup Funds from The University of Newcastle, Australia.

\bibliography{aaai2026_Arxiv}

@inproceedings{yan2023feature,
	title={Feature prediction diffusion model for video anomaly detection},
	author={Yan, Cheng and Zhang, Shiyu and Liu, Yang and Pang, Guansong and Wang, Wenjun},
	booktitle={ICCV},
	year={2023}
}

@inproceedings{wang2023openoccupancy,
	title={Openoccupancy: A large scale benchmark for surrounding semantic occupancy perception},
	author={Wang, Xiaofeng and Zhu, Zheng and Xu, Wenbo and Zhang, Yunpeng and Wei, Yi and Chi, Xu and Ye, Yun and Du, Dalong and Lu, Jiwen and Wang, Xingang},
	booktitle={ICCV},
	year={2023}
}

@inproceedings{thatipelli2022spatio,
	title={Spatio-temporal relation modeling for few-shot action recognition},
	author={Thatipelli, Anirudh and Narayan, Sanath and Khan, Salman and Anwer, Rao Muhammad and Khan, Fahad Shahbaz and Ghanem, Bernard},
	booktitle={CVPR},
	year={2022}
}

@inproceedings{xing2023revisiting,
	title={Revisiting the Spatial and Temporal Modeling for Few-shot Action Recognition},
	author={Xing, Jiazheng and Wang, Mengmeng and Liu, Yong and Mu, Boyu},
	booktitle={AAAI},
	year={2023}
}

@inproceedings{zhang2023importance,
	title={On the Importance of Spatial Relations for Few-shot Action Recognition},
	author={Zhang, Yilun and Fu, Yuqian and Ma, Xingjun and Qi, Lizhe and Chen, Jingjing and Wu, Zuxuan and Jiang, Yu-Gang},
	booktitle={ACM MM},
	year={2023}
}

@inproceedings{yu2023multi,
	title={Multi-Speed Global Contextual Subspace Matching for Few-Shot Action Recognition},
	author={Yu, Tianwei and Chen, Peng and Dang, Yuanjie and Huan, Ruohong and Liang, Ronghua},
	booktitle={ACM MM},
	year={2023}
}

@inproceedings{xing2023boosting,
	title={Boosting Few-shot Action Recognition with Graph-guided Hybrid Matching},
	author={Xing, Jiazheng and Wang, Mengmeng and Ruan, Yudi and Chen, Bofan and Guo, Yaowei and Mu, Boyu and Dai, Guang and Wang, Jingdong and Liu, Yong},
	booktitle={ICCV},
	year={2023}
}

@inproceedings{wanyan2023active,
	title={Active Exploration of Multimodal Complementarity for Few-Shot Action Recognition},
	author={Wanyan, Yuyang and Yang, Xiaoshan and Chen, Chaofan and Xu, Changsheng},
	booktitle={CVPR},
	year={2023}
}

@inproceedings{liu2023lite,
	title={Lite-MKD: A Multi-modal Knowledge Distillation Framework for Lightweight Few-shot Action Recognition},
	author={Liu, Baolong and Zheng, Tianyi and Zheng, Peng and Liu, Daizong and Qu, Xiaoye and Gao, Junyu and Dong, Jianfeng and Wang, Xun},
	booktitle={ACM MM},
	year={2023}
}

@inproceedings{perrett2021temporal,
	title={Temporal-relational crosstransformers for few-shot action recognition},
	author={Perrett, Toby and Masullo, Alessandro and Burghardt, Tilo and Mirmehdi, Majid and Damen, Dima},
	booktitle={CVPR},
	year={2021}
}

@inproceedings{wang2023molo,
	title={MoLo: Motion-augmented Long-short Contrastive Learning for Few-shot Action Recognition},
	author={Wang, Xiang and Zhang, Shiwei and Qing, Zhiwu and Gao, Changxin and Zhang, Yingya and Zhao, Deli and Sang, Nong},
	booktitle={CVPR},
	year={2023}
}

@inproceedings{wu2022motion,
	title={Motion-modulated temporal fragment alignment network for few-shot action recognition},
	author={Wu, Jiamin and Zhang, Tianzhu and Zhang, Zhe and Wu, Feng and Zhang, Yongdong},
	booktitle={CVPR},
	year={2022}
}

@inproceedings{wang2022hybrid,
	title={Hybrid relation guided set matching for few-shot action recognition},
	author={Wang, Xiang and Zhang, Shiwei and Qing, Zhiwu and Tang, Mingqian and Zuo, Zhengrong and Gao, Changxin and Jin, Rong and Sang, Nong},
	booktitle={CVPR},
	year={2022}
}

@inproceedings{fu2020depth,
	title={Depth guided adaptive meta-fusion network for few-shot video recognition},
	author={Fu, Yuqian and Zhang, Li and Wang, Junke and Fu, Yanwei and Jiang, Yu-Gang},
	booktitle={ACM MM},
	year={2020}
}

@inproceedings{huang2024soap,
	title={SOAP: Enhancing Spatio-Temporal Relation and Motion Information Capturing for Few-Shot Action Recognition},
	author={Huang, Wenbo and Zhang, Jinghui and Qian, Xuwei and Wu, Zhen and Wang, Meng and Zhang, Lei},
	booktitle={ACM MM},
	year={2024}
}

@inproceedings{huang2025manta,
	title={Manta: Enhancing Mamba for Few-Shot Action Recognition of Long Sub-Sequence},
	author={Huang, Wenbo and Zhang, Jinghui and Li, Guang and Zhang, Lei and Wang, Shuoyuan and Dong, Fang and Jin, Jiahui and Ogawa, Takahiro and Haseyama, Miki},
	booktitle={AAAI},
	year={2025}
}

@inproceedings{peng2023rwkv,
	title={Rwkv: Reinventing rnns for the transformer era},
	author={Peng, Bo and Alcaide, Eric and Anthony, Quentin and Albalak, Alon and Arcadinho, Samuel and Biderman, Stella and Cao, Huanqi and Cheng, Xin and Chung, Michael and Grella, Matteo and others},
	booktitle={EMNLP},
	year={2023}
}

@inproceedings{peng2024eagle,
	title={Eagle and finch: Rwkv with matrix-valued states and dynamic recurrence},
	author={Peng, Bo and Goldstein, Daniel and Anthony, Quentin and Albalak, Alon and Alcaide, Eric and Biderman, Stella and Cheah, Eugene and Du, Xingjian and Ferdinan, Teddy and Hou, Haowen and others},
	booktitle={COLM},
	year={2024}
}

@inproceedings{cao2020few,
	title={Few-shot video classification via temporal alignment},
	author={Cao, Kaidi and Ji, Jingwei and Cao, Zhangjie and Chang, Chien-Yi and Niebles, Juan Carlos},
	booktitle={CVPR},
	year={2020}
}

@article{fei2006one,
	title={One-shot learning of object categories},
	author={Fei-Fei, Li and Fergus, Robert and Perona, Pietro},
	journal={IEEE Transactions on Pattern Analysis and Machine Intelligence},
	year={2006},
	publisher={IEEE}
}

@inproceedings{chen2019image,
	title={Image deformation meta-networks for one-shot learning},
	author={Chen, Zitian and Fu, Yanwei and Wang, Yu-Xiong and Ma, Lin and Liu, Wei and Hebert, Martial},
	booktitle={CVPR},
	year={2019}
}

@inproceedings{hariharan2017low,
	title={Low-shot visual recognition by shrinking and hallucinating features},
	author={Hariharan, Bharath and Girshick, Ross},
	booktitle={ICCV},
	year={2017}
}

@inproceedings{wang2018low,
	title={Low-shot learning from imaginary data},
	author={Wang, Yu-Xiong and Girshick, Ross and Hebert, Martial and Hariharan, Bharath},
	booktitle={CVPR},
	year={2018}
}

@inproceedings{li2020adversarial,
	title={Adversarial feature hallucination networks for few-shot learning},
	author={Li, Kai and Zhang, Yulun and Li, Kunpeng and Fu, Yun},
	booktitle={CVPR},
	year={2020}
}

@inproceedings{zhang2018metagan,
	title={Metagan: An adversarial approach to few-shot learning},
	author={Zhang, Ruixiang and Che, Tong and Ghahramani, Zoubin and Bengio, Yoshua and Song, Yangqiu},
	booktitle={NeurIPS},
	year={2018}
}

@inproceedings{finn2017model,
	title={Model-agnostic meta-learning for fast adaptation of deep networks},
	author={Finn, Chelsea and Abbeel, Pieter and Levine, Sergey},
	booktitle={ICML},
	year={2017}
}

@inproceedings{jamal2019task,
	title={Task agnostic meta-learning for few-shot learning},
	author={Jamal, Muhammad Abdullah and Qi, Guo-Jun},
	booktitle={CVPR},
	year={2019}
}

@inproceedings{ravi2017optimization,
	title={Optimization as a model for few-shot learning},
	author={Ravi, Sachin and Larochelle, Hugo},
	booktitle={ICLR},
	year={2017}
}

@inproceedings{rusu2018meta,
	title={Meta-learning with latent embedding optimization},
	author={Rusu, Andrei A and Rao, Dushyant and Sygnowski, Jakub and Vinyals, Oriol and Pascanu, Razvan and Osindero, Simon and Hadsell, Raia},
	booktitle={ICLR},
	year={2018}
}

@inproceedings{rajeswaran2019meta,
	title={Meta-learning with implicit gradients},
	author={Rajeswaran, Aravind and Finn, Chelsea and Kakade, Sham M and Levine, Sergey},
	booktitle={NeurIPS},
	year={2019}
}

@inproceedings{snell2017prototypical,
	title={Prototypical networks for few-shot learning},
	author={Snell, Jake and Swersky, Kevin and Zemel, Richard},
	booktitle={NeurIPS},
	year={2017}
}

@inproceedings{oreshkin2018tadam,
	title={Tadam: Task dependent adaptive metric for improved few-shot learning},
	author={Oreshkin, Boris and Rodr{\'\i}guez L{\'o}pez, Pau and Lacoste, Alexandre},
	booktitle={NeurIPS},
	year={2018}
}

@inproceedings{hao2019collect,
	title={Collect and select: Semantic alignment metric learning for few-shot learning},
	author={Hao, Fusheng and He, Fengxiang and Cheng, Jun and Wang, Lei and Cao, Jianzhong and Tao, Dacheng},
	booktitle={CVPR},
	year={2019}
}

@inproceedings{sung2018learning,
	title={Learning to compare: Relation network for few-shot learning},
	author={Sung, Flood and Yang, Yongxin and Zhang, Li and Xiang, Tao and Torr, Philip HS and Hospedales, Timothy M},
	booktitle={CVPR},
	year={2018}
}

@inproceedings{wang2020cooperative,
	title={Cooperative bi-path metric for few-shot learning},
	author={Wang, Zeyuan and Zhao, Yifan and Li, Jia and Tian, Yonghong},
	booktitle={ACM MM},
	year={2020}
}

@inproceedings{zhang2021learning,
	title={Learning implicit temporal alignment for few-shot video classification},
	author={Zhang, Songyang and Zhou, Jiale and He, Xuming},
	booktitle={IJCAI},
	year={2021}
}

@inproceedings{li2022ta2n,
	title={Ta2n: Two-stage action alignment network for few-shot action recognition},
	author={Li, Shuyuan and Liu, Huabin and Qian, Rui and Li, Yuxi and See, John and Fei, Mengjuan and Yu, Xiaoyuan and Lin, Weiyao},
	booktitle={AAAI},
	year={2022}
}

@article{duan2024vision,
	title={Vision-rwkv: Efficient and scalable visual perception with rwkv-like architectures},
	author={Duan, Yuchen and Wang, Weiyun and Chen, Zhe and Zhu, Xizhou and Lu, Lewei and Lu, Tong and Qiao, Yu and Li, Hongsheng and Dai, Jifeng and Wang, Wenhai},
	journal={arXiv preprint arXiv:2403.02308},
	year={2024}
}

@article{fei2024diffusion,
	title={Diffusion-rwkv: Scaling rwkv-like architectures for diffusion models},
	author={Fei, Zhengcong and Fan, Mingyuan and Yu, Changqian and Li, Debang and Huang, Junshi},
	journal={arXiv preprint arXiv:2404.04478},
	year={2024}
}

@article{gu2024rwkv,
	title={RWKV-CLIP: A Robust Vision-Language Representation Learner},
	author={Gu, Tiancheng and Yang, Kaicheng and An, Xiang and Feng, Ziyong and Liu, Dongnan and Cai, Weidong and Deng, Jiankang},
	journal={arXiv preprint arXiv:2406.06973},
	year={2024}
}

@article{he2024pointrwkv,
	title={Pointrwkv: Efficient rwkv-like model for hierarchical point cloud learning},
	author={He, Qingdong and Zhang, Jiangning and Peng, Jinlong and He, Haoyang and Li, Xiangtai and Wang, Yabiao and Wang, Chengjie},
	journal={arXiv preprint arXiv:2405.15214},
	year={2024}
}

@article{yuan2024mamba,
	title={Mamba or rwkv: Exploring high-quality and high-efficiency segment anything model},
	author={Yuan, Haobo and Li, Xiangtai and Qi, Lu and Zhang, Tao and Yang, Ming-Hsuan and Yan, Shuicheng and Loy, Chen Change},
	journal={arXiv preprint arXiv:2406.19369},
	year={2024}
}

@inproceedings{zhao2022c3,
	title={C3-stisr: Scene text image super-resolution with triple clues},
	author={Zhao, Minyi and Wang, Miao and Bai, Fan and Li, Bingjia and Wang, Jie and Zhou, Shuigeng},
	booktitle={IJCAI},
	year={2022}
}

@inproceedings{goyal2017something,
	title={The" something something" video database for learning and evaluating visual common sense},
	author={Goyal, Raghav and Ebrahimi Kahou, Samira and Michalski, Vincent and Materzynska, Joanna and Westphal, Susanne and Kim, Heuna and Haenel, Valentin and Fruend, Ingo and Yianilos, Peter and Mueller-Freitag, Moritz and others},
	booktitle={ICCV},
	year={2017}
}

@inproceedings{carreira2017quo,
	title={Quo vadis, action recognition? a new model and the kinetics dataset},
	author={Carreira, Joao and Zisserman, Andrew},
	booktitle={CVPR},
	year={2017}
}

@article{kay2017kinetics,
	title={The kinetics human action video dataset},
	author={Kay, Will and Carreira, Joao and Simonyan, Karen and Zhang, Brian and Hillier, Chloe and Vijayanarasimhan, Sudheendra and Viola, Fabio and Green, Tim and Back, Trevor and Natsev, Paul and others},
	journal={arXiv preprint arXiv:1705.06950},
	year={2017}
}

@inproceedings{kuehne2011hmdb,
	title={HMDB: a large video database for human motion recognition},
	author={Kuehne, Hildegard and Jhuang, Hueihan and Garrote, Est{\'\i}baliz and Poggio, Tomaso and Serre, Thomas},
	booktitle={ICCV},
	year={2011}
}

@inproceedings{zhang2020few,
	title={Few-shot action recognition with permutation-invariant attention},
	author={Zhang, Hongguang and Zhang, Li and Qi, Xiaojuan and Li, Hongdong and Torr, Philip HS and Koniusz, Piotr},
	booktitle={ECCV},
	year={2020}
}

@inproceedings{zhu2018compound,
	title={Compound memory networks for few-shot video classification},
	author={Zhu, Linchao and Yang, Yi},
	booktitle={ECCV},
	year={2018}
}

@inproceedings{wang2016temporal,
	title={Temporal segment networks: Towards good practices for deep action recognition},
	author={Wang, Limin and Xiong, Yuanjun and Wang, Zhe and Qiao, Yu and Lin, Dahua and Tang, Xiaoou and Van Gool, Luc},
	booktitle={ECCV},
	year={2016}
}

@inproceedings{zhang2024continuous,
	title={Continuous-multiple image outpainting in one-step via positional query and a diffusion-based approach},
	author={Zhang, Shaofeng and Huang, Jinfa and Zhou, Qiang and Wang, Zhibin and Wang, Fan and Luo, Jiebo and Yan, Junchi},
	booktitle={ICLR},
	year={2024}
}

@article{van2008visualizing,
	title={Visualizing data using t-SNE.},
	author={Van der Maaten, Laurens and Hinton, Geoffrey},
	journal={Journal of machine learning research},
	year={2008}
}

@article{li2024benchmarking,
	title={Benchmarking Badminton Action Recognition with a New Fine-Grained Dataset},
	author={Li, Qi and Chiu, Tzu-Chen and Huang, Hsiang-Wei and Sun, Min-Te and Ku, Wei-Shinn},
	journal={arXiv preprint arXiv:2403.12385},
	year={2024}
}

@article{liao2023deep,
	title={Deep learning for camera calibration and beyond: A survey},
	author={Liao, Kang and Nie, Lang and Huang, Shujuan and Lin, Chunyu and Zhang, Jing and Zhao, Yao and Gabbouj, Moncef and Tao, Dacheng},
	journal={arXiv preprint arXiv:2303.10559},
	year={2023}
}

@article{lai2021correcting,
	title={Correcting face distortion in wide-angle videos},
	author={Lai, Wei-Sheng and Shih, Yichang and Liang, Chia-Kai and Yang, Ming-Hsuan},
	journal={IEEE Transactions on Image Processing},
	year={2021},
	publisher={IEEE}
}

@article{hold2023perceptual,
	title={A perceptual measure for deep single image camera and lens calibration},
	author={Hold-Geoffroy, Yannick and Pich{\'e}-Meunier, Dominique and Sunkavalli, Kalyan and Bazin, Jean-Charles and Rameau, Fran{\c{c}}ois and Lalonde, Jean-Fran{\c{c}}ois},
	journal={IEEE Transactions on Pattern Analysis and Machine Intelligence},
	year={2023},
	publisher={IEEE}
}

@inproceedings{zhang2025madcow,
	title={MaDCoW: Marginal Distortion Correction for Wide-Angle Photography with Arbitrary Objects},
	author={Zhang, Kevin and Huang, Jia-Bin and Echevarria, Jose and DiVerdi, Stephen and Hertzmann, Aaron},
	booktitle={CVPR},
	year={2025}
}

@inproceedings{lee2021ctrl,
	title={Ctrl-c: Camera calibration transformer with line-classification},
	author={Lee, Jinwoo and Go, Hyunsung and Lee, Hyunjoon and Cho, Sunghyun and Sung, Minhyuk and Kim, Junho},
	booktitle={ICCV},
	year={2021}
}

@article{hu2022miniature,
	title={Miniature optoelectronic compound eye camera},
	author={Hu, Zhi-Yong and Zhang, Yong-Lai and Pan, Chong and Dou, Jian-Yu and Li, Zhen-Ze and Tian, Zhen-Nan and Mao, Jiang-Wei and Chen, Qi-Dai and Sun, Hong-Bo},
	year={2022},
	journal={Nature Communications},
	publisher={Nature Publishing Group}
}
\end{document}